\documentclass[11pt]{article}

\IfFileExists{acl-style-files/acl.sty}{%
  \usepackage[final]{acl-style-files/acl}%
}{%
  \usepackage[final]{acl}%
}

\usepackage{times}
\usepackage{latexsym}

\usepackage[T1]{fontenc}

\usepackage[utf8]{inputenc}

\usepackage{microtype}

\usepackage{graphicx}
\usepackage{algorithm}
\usepackage{algpseudocode}
\usepackage{enumitem}
\usepackage{bbm}
\usepackage{makecell}
\usepackage{amsmath}
\usepackage{amssymb}
\usepackage{booktabs}
\usepackage{multirow}
\usepackage{pifont}
\usepackage{microtype}
\usepackage{graphicx}
\usepackage{subcaption}
\usepackage{booktabs} 
\usepackage{changepage}  
\usepackage{subcaption}
\usepackage{svg}
\usepackage[table]{xcolor}
\usepackage{colortbl}
\usepackage{array}
\usepackage{tabularx}
\usepackage{subcaption}
\usepackage{booktabs}
\usepackage{xcolor}

\usepackage[most]{tcolorbox}
\IfFileExists{fontawesome5.sty}{\usepackage{fontawesome5}}{\newcommand{\faBook}{}} 

\tcbset{
  promptstyle/.style={
    enhanced,
    breakable,
    colback=green!6,
    colframe=green!55!black,
    arc=2mm,
    boxrule=0.8pt,
    left=6pt,right=6pt,top=6pt,bottom=6pt,
    colbacktitle=green!55!black,
    coltitle=white,
    fonttitle=\bfseries,
    boxed title style={arc=2mm, boxrule=0pt, left=4pt, right=4pt},
    attach boxed title to top left={xshift=0mm,yshift=-1mm},
  }
}

\newtcolorbox{promptbox}[1]{promptstyle, title=\faBook\quad #1}

\newtcolorbox{promptboxplain}[1]{promptstyle, title=#1} 
\usepackage{enumitem}

\tcbuselibrary{listings,breakable,skins}
\usepackage{listings}

\usepackage[T1]{fontenc}
\usepackage[dvipsnames,table]{xcolor} 
\definecolor{mygrey}{gray}{0.4}       
\tcbset{
  promptcodeframe/.style={
    enhanced,
    breakable,
    sharp corners,
    colframe=ForestGreen,
    colback=white,
    boxrule=3pt,
    boxsep=0.5pt,
    left=6pt,right=6pt,top=6pt,bottom=6pt,
    shadow={3pt}{-3pt}{0pt}{opacity=1,mygrey},
  }
}

\newtcblisting{promptboxcode}[1]{
  promptcodeframe,
  title={#1},
  listing only,
  listing options={
    basicstyle=\ttfamily\scriptsize,
    breaklines=true,
    breakatwhitespace=false,
    showstringspaces=false,
    columns=fullflexible,
    xleftmargin=0pt,
    xrightmargin=0pt
  }
}
\usepackage[most]{tcolorbox}

\usepackage{listings}
\usepackage{makecell}
\usepackage{multirow}
\usepackage{amsmath}
\usepackage{amssymb}
\usepackage{mathtools}
\usepackage{amsthm}
\usepackage[table]{xcolor} 
\usepackage{booktabs}
\usepackage{pifont}   
\newcommand{\drop}[1]{\textcolor{red}{\scriptsize\,($\downarrow$#1)}}
\newcolumntype{Y}{>{\centering\arraybackslash}X}
\definecolor{HFSearchBlue}{RGB}{247,252,255}
\definecolor{HFTrainBlue}{RGB}{232,242,255}
\definecolor{HFForgePurple}{RGB}{247,245,254}
\DeclareRobustCommand{\legendbox}[2]{\begingroup\setlength{\fboxsep}{1pt}\colorbox{#1}{#2}\endgroup}
\title{HarnessForge: Joint Harness and Policy Evolution for Adaptive Agent Systems}

\author{
 \textbf{Mingju Chen\textsuperscript{1}},
 \textbf{Can Lv\textsuperscript{1}},
 \textbf{Guibin Zhang},
 \textbf{Heng Chang\textsuperscript{2}},
 \textbf{Shiji Zhou\textsuperscript{1}}
\\
\\
 \textsuperscript{1}Beijing Advanced Innovation Center for Future Blockchain and Privacy Computing, \\ School of Artificial Intelligence, Beihang University,
 \textsuperscript{2}Tsinghua University
\\
 \small{
   \textbf{Project Lead:} Heng Chang,
   \textbf{Corresponding to:} Shiji Zhou <\href{mailto:email@domain}{zhoushiji25@buaa.edu.cn}>
 }
}

\begin{document}
\maketitle
\begin{abstract}
LLM agents are increasingly expected to operate across heterogeneous task regimes that require distinct execution paradigms. This challenges fixed agent systems and motivates system-level meta-adaptation beyond isolated component updates. While existing works have adapted external harness or trained underlying reasoning policies, full-system adaptation remains insufficiently characterized. The adaptation space between structure and execution is rarely made explicit, and the compatibility between the external harness and the internal reasoner is not optimized jointly.
We propose HarnessForge, a meta-adaptive framework for evolving LLM agent systems. HarnessForge formulates an agent system as a harness--policy pair, defining a stable adaptation space that separates harness-level execution structure from policy-level reasoning behavior. It then performs harness--policy co-evolution through fault-guided harness tailoring and harness-conditioned policy alignment.
Experiments across five benchmarks from diverse domains show that HarnessForge consistently improves both Qwen3-4B and Qwen3-8B backbones, outperforming harness-only and policy-only baselines with gains of up to 12.0\% over the strongest baseline and achieving favorable rollout-efficiency tradeoffs, demonstrating that harness--policy co-evolution is effective, and that executable compatibility between the  harness and reasoning policy is essential for agent-system adaptation. The code is available at \url{https://github.com/mingju-c/HarnessForge}.
\end{abstract}

\begin{figure}[t]
  \begin{center}
    \centerline{\includegraphics[width=0.99\columnwidth]{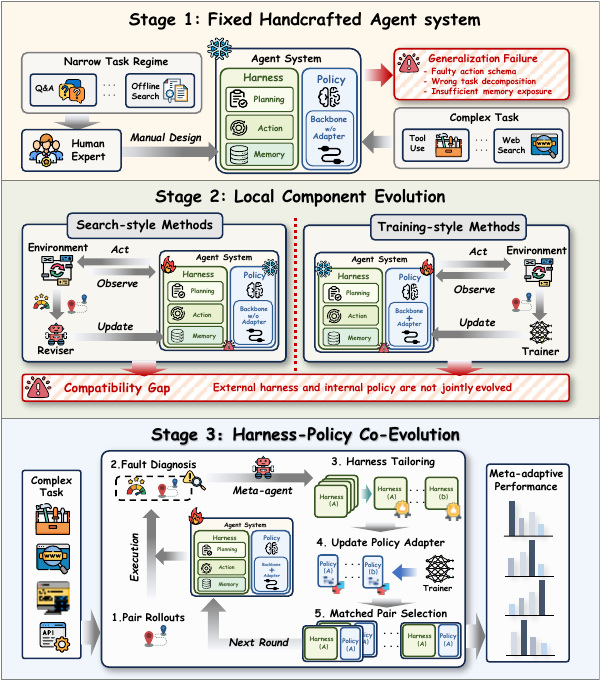}}
    \caption{A three-stage view of LLM agent adaptation: fixed handcrafted systems, local component adaptation, and harness--policy co-evolution (HarnessForge).}
    \label{fig:compare}
  \end{center}
  \vskip -0.4 in
\end{figure}
\section{Introduction}

LLM agents are increasingly deployed in complex and heterogeneous task regimes, including multi-step tool use \citep{schick2023toolformer,qian2026toolrl}, retrieval-heavy reasoning \citep{jin2025searchr1trainingllmsreason,shen2025satorireinforcementlearningchainofactionthought}, web interaction \citep{NEURIPS2022_82ad13ec}, and stateful multi-turn tasks \citep{yao2024taubenchbenchmarktoolagentuserinteraction,shinn2023reflexion}. These regimes differ not only in task difficulty, but also in the structural requirements they impose on agent execution. Some require explicit task decomposition and verification, some rely on strict action schemas and tool-use protocols, while others demand persistent memory exposure, or state tracking. Such diversity suggests that there is unlikely to be a single fixed agent system that performs optimally across regimes. Instead, agent systems should be able to meta-adapt their execution paradigms to the target task regime.

As shown in Fig.~\ref{fig:compare}, recent work has begun to move toward this goal by treating agent components as adaptive objects rather than fixed hand-written artifacts \citep{ICLR2025_36b7acf6,zhang2025aflow,shang2025agentsquare,zhang2025multiagent}. Concretely, some search-style methods optimize external execution structures, including workflows, tool-use procedures, role assignments, memory management, or execution graphs \citep{wu2024autogen,hong2024metagpt,zhong2023memorybankenhancinglargelanguage,packer2024memgptllmsoperatingsystems,zhang2025memevolvemetaevolutionagentmemory}, showing that external execution structures can be searched, revised, or evolved. Other training-style methods adapt the model policy through supervised learning, preference learning, or reinforcement learning on agentic trajectories \citep{shinn2023reflexion,madaan2023selfrefine,qian2026toolrl,li2025torlscalingtoolintegratedrl,shao2024deepseekmathpushinglimitsmathematical}, improving the model's internal execution behavior. These works demonstrate that different components of an agent system can be adapted.

However, this component-level view remains insufficient for system-level meta-adaptation. The core issue lies in the \textbf{\emph{adaptation target}}: existing methods typically optimize external harness or internal policies as separate objects, whereas a full LLM agent system operates as a coupled harness--policy pair. This coupling is especially salient in resource-constrained settings, where the harness provides structural support for limited model capabilities and the policy must learn to execute the harness-induced execution paradigm. Moreover, existing adaptation still faces a \textbf{\emph{compatibility gap}}. A more expressive harness may expose useful planning, action, or memory structures, yet fail if the reasoner cannot reliably execute them; conversely, a stronger policy may still be constrained by a harness that exposes unsuitable states, actions, or control signals. Effective system-level adaptation should therefore go beyond optimizing either side alone and instead co-evolve the external harness and internal policy to improve their compatibility.

Motivated by this view, we formulate an LLM agent system as a harness--policy pair, making the coupled harness and policy the explicit unit of adaptation. The harness specifies the external execution interface, including planning, action, and memory structures that shape agent behavior; the policy captures how the reasoner executes within this interface. Based on this formulation, we propose HarnessForge, a framework for harness--policy co-adaptation. On the harness side, HarnessForge uses rollout diagnostics with a meta-agent to perform fault-guided harness tailoring over planning, action, and memory components. On the policy side, it trains harness-conditioned adapters from curated trajectories to align the reasoner with the selected harness. Finally, HarnessForge selects and evolves harness--policy pairs, optimizing task-regime-adapted agent systems rather than isolated workflows or policies across evolutionary rounds.

Our contributions are summarized as follows:
\begin{itemize}[leftmargin=1.35em, labelsep=0.45em, itemsep=1pt, topsep=2pt, parsep=0pt]
    \item We reformulate system-level LLM agent adaptation from local component optimization to harness--policy pair evolution, treating the coupled external harness and internal policy as the basic unit of optimization.
    
    \item We propose HarnessForge, a meta-adaptive co-evolution framework that coordinates fault-guided harness tailoring with harness-conditioned policy alignment to improve pair-level executable compatibility.
    
    \item We evaluate HarnessForge across five benchmarks and two backbones, improving over the strongest harness-only and policy-only baselines by 3.56\% on average and up to 12.0\%, while preserving favorable rollout–performance trade offs and harness–policy compatibility.
\end{itemize}

\begin{figure*}[ht]
  \begin{center}
    \centerline{\includegraphics[width=2.1\columnwidth]{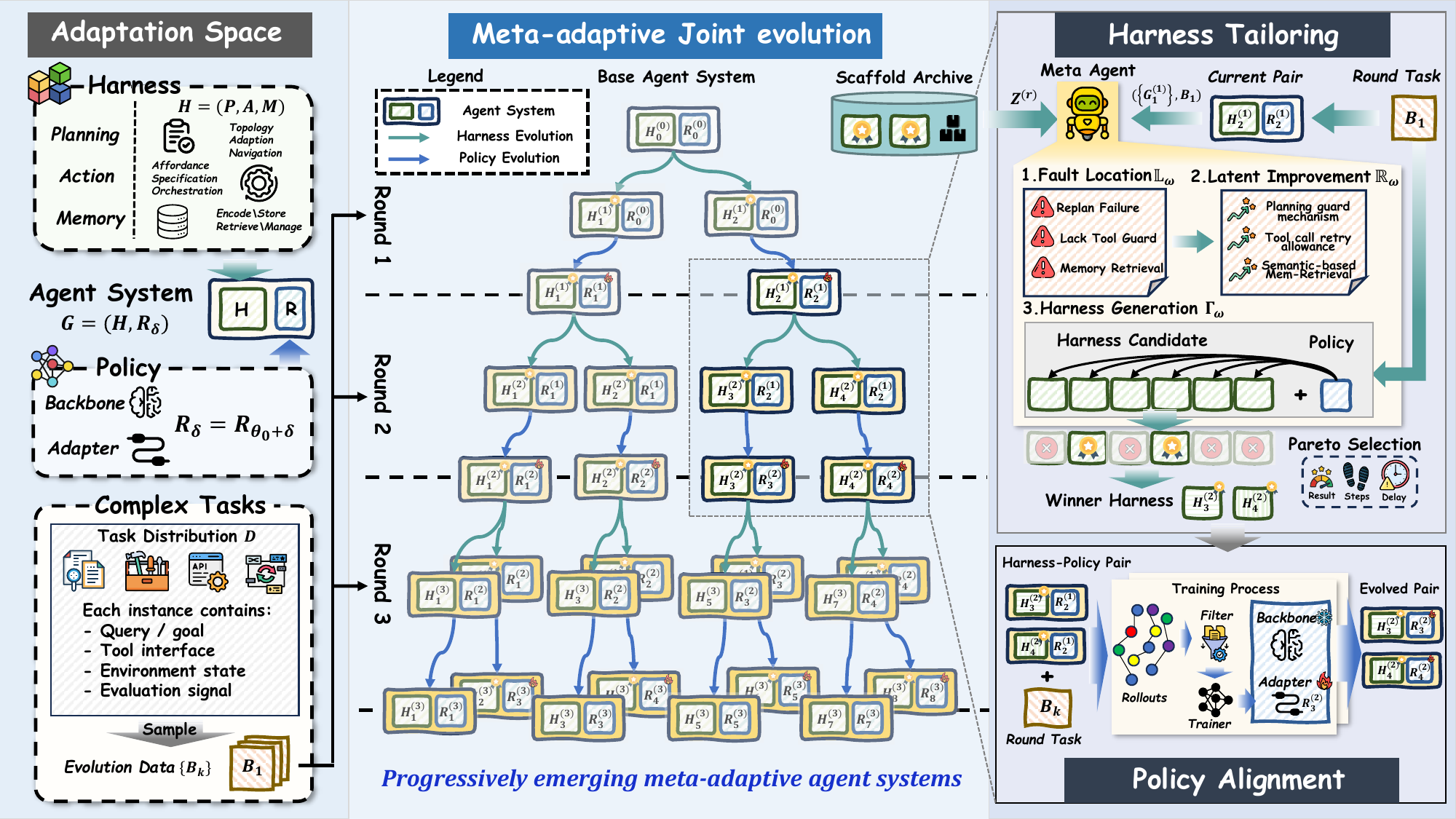}}
    \caption{Overview of the HarnessForge co-evolution workflow. Starting from a harness--policy pair, each round diagnoses execution failures, tailors the harness over planning, action, and memory modules, trains a harness-conditioned adapter from curated trajectories, and selects improved matched pairs for the next round.}
    \label{fig:workflow}
  \end{center}
  \vskip -0.4in
\end{figure*}

\section{Related Work}

\paragraph{Optimization for Agent System Design}

Prior work shows that LLM-agent capabilities are strongly shaped by the external harness used to organize reasoning and execution.
Early prompting and interaction paradigms, such as Chain-of-Thought \citep{wei2022chain}, Plan-and-Solve \citep{wang-etal-2023-plan}, and ReAct \citep{yao2023react}, introduce explicit reasoning traces, plans, actions, and observations.
Agent frameworks further expose more components, including roles, protocols, and memory modules in AutoGen \citep{wu2024autogen}, MetaGPT \citep{hong2024metagpt}, and memory-augmented agent systems \citep{zhong2023memorybankenhancinglargelanguage,packer2024memgptllmsoperatingsystems,zhang2025memevolvemetaevolutionagentmemory}.
More recent search-style methods automate harness design through workflow or architecture search, including ADAS \citep{ICLR2025_36b7acf6}, AFlow \citep{zhang2025aflow}, AgentSquare \citep{shang2025agentsquare}, MaAS \citep{zhang2025multiagent}, AutoHarness \citep{lou2026autoharnessimprovingllmagents}, Meta-Harness \citep{lee2026metaharnessendtoendoptimizationmodel} and MermaidFlow \citep{zheng2025mermaidflowredefiningagenticworkflow}.
These works reduce manual engineering and establish harnesses as important optimization targets, but they mainly optimize external structures while leaving compatibility with the model-side executor implicit.
\paragraph{Agentic RL for Policy Evolution}
Another line of work optimizes the model-side policy. Recent agentic RL methods train models from interactive trajectories with task rewards, tool feedback, or environment signals. For search and tool use, Search-R1 \citep{jin2025searchr1trainingllmsreason}, ToolRL \citep{qian2026toolrl}, and ToRL \citep{li2025torlscalingtoolintegratedrl} optimize when and how models issue external actions. For long-horizon agent execution, GiGPO \citep{NEURIPS2025_420c9f77}, TreeRL \citep{hou-etal-2025-treerl}, and ARPO \citep{dong2026agentic} address credit assignment and trajectory-level optimization, while Planner-R1 \citep{zhu2025plannerr1rewardshapingenables} and Memory-R1 \citep{yan2026memoryr1enhancinglargelanguage} extend RL to planning and memory management. Broader reasoning-RL systems, including DeepseekMATH (GRPO) \citep{shao2024deepseekmathpushinglimitsmathematical}, DAPO \citep{NEURIPS2025_a4277440}, GSPO \citep{zheng2025groupsequencepolicyoptimization}, Satori \citep{shen2025satorireinforcementlearningchainofactionthought}, and Absolute Zero \citep{zhao2025absolutezeroreinforcedselfplay}, further show that RL can improve reasoning, exploration, and self-generated curricula. These methods strengthen the internal executor, but typically assume a fixed, externally specified interaction loop or task interface.

\vspace{-2mm}
\section{Methodology}
\label{sec:methodology}

Fig.~\ref{fig:workflow} shows the overall workflow of HarnessForge. Sec.~\ref{sec:Preliminary} formalizes the agent system, and evaluation criteria. Sec.~\ref{sec:meta-adaptive-coevolution} presents our meta-adaptive joint evolution mechanism. Secs.~\ref{sec:fault-guided-harness-tailoring} and~\ref{sec:trajectory-guided-reasoning-evolution} detail harness and policy evolution, respectively.
\vspace{-2mm}
\subsection{Preliminary}
\label{sec:Preliminary}

\paragraph{Agent-System Formulation.}
We formulate an LLM agent system \(\mathcal{G}\) as the coupling of an external harness \(\mathcal{H}\) and an adapted reasoner \(\mathcal{R}_{\delta}\):
\begin{equation}
    \mathcal{G}
    =
    (\mathcal{H},\mathcal{R}_{\delta}),
    \mathcal{H}=(\mathcal{P},\mathcal{A},\mathcal{M}),
    \mathcal{R}_{\delta}
    =
    \mathcal{R}_{\theta_0+\delta}.
\label{eq:agent-individual}
\end{equation}
Here \(\mathcal{H}\) is the editable execution harness, which decomposes into three execution-layer components: \(\mathcal{P}\) denotes the planning component, including task decomposition, replanning, and termination. \(\mathcal{A}\) denotes the action component, including tool interfaces, role assignment, and orchestration rules. \(\mathcal{M}\) denotes the memory component, including what is written, retrieved, summarized, and exposed to future decisions. The adapted reasoner \(\mathcal{R}_{\delta}\) denotes the reasoning component, which parameterizes the policy that executes under this harness, and \(\delta\) is a lightweight adapter on base reasoner \(\mathcal{R}_{\theta_0}\).

\paragraph{Evaluation Criteria.}
Given a task \(x\) and trajectory \(\tau_x\), we let \(\boldsymbol{\phi}(\tau,x)\) collect final response quality within environment, negative token cost, negative latency to evaluate the agent system, with larger values preferred in every dimension. For a batch \(B\subset\mathcal{D}\), we define fitness indicator $\mathbf{J}(\mathcal{G};B)$:
\begin{equation}
    \mathbf{J}(\mathcal{G};B)
    =
    \frac{1}{|B|}
    \sum_{x\in B}
    \boldsymbol{\phi}\big(\tau_x(\mathcal{G}),x\big).
\label{eq:batch-fitness}
\end{equation}
These criteria are used for Pareto-based validation and selection of candidate systems; the detailed evaluation procedure is provided in App.~\ref{app:Evaluator}.

\subsection{Meta-Adaptive Joint Evolution}
\label{sec:meta-adaptive-coevolution}

Previous work typically improves only individual components of an agent system and overlooks the compatibility between the external harness and the internal reasoning policy.
To remedy this gap, HarnessForge adopts a meta-adaptive joint evolution mechanism over iterative rounds. The two evolution processes are mutually reinforcing: \textit{better harnesses induce more structured and informative trajectories, while stronger reasoning policies execute harness protocols more faithfully.}

\paragraph{Round Structure.}

At round \(r\), HarnessForge maintains a population of the agent systems $\mathbb{G}^{(r)}$: 
\begin{equation}
\mathbb{G}^{(r)}
=
\{
\mathcal{G}_i^{(r)} = (\mathcal{H}_i^{(r)},\mathcal{R}^{(r)}_{\delta_i})
\}_{i\in I^{(r)}},
\label{eq:population}
\end{equation}
where each element is an executable harness--policy pair.
The initial round
\(
r=0
\)
starts from a singleton population containing a manually designed base harness and the frozen base reasoner.

Given the evolution batch data \(B_r\), HarnessForge first performs harness tailoring. For each agent, $\mathcal{G}_i^{(r)}$ executes tasks on \(B_r\) and collects trajectories, execution statistics, and environment feedback.
A meta-agent tailoring operator \(T_{\psi}\) then updates the harness population through controlled executable harness editing, including fault attribution, archive-guided improvement, candidate generation, and budgeted Pareto selection:
\begin{equation}
\mathcal{C}^{(r+1)}
=
T_{\psi}\big(\{\mathcal{G}_i^{(r)}\}_{i\in I^{(r)}},\mathcal{Z}^{(r)},B_r\big),
\label{eq:harness-selection-overview}
\end{equation}
where \(\mathcal{C}^{(r+1)}\) is the survivor harness set retained for the next policy-evolution stage.

Conditioned on the survivor harness set \(\mathcal{C}^{(r+1)}\),
HarnessForge performs policy alignment for each harness through a policy-evolution operator $E_{\eta}$:
\begin{equation}
\mathcal{R}^{(r+1)}_{\delta_k}=E_{\eta}\big((\mathcal{H}_k^{(r+1)},\mathcal{R}^{(r)}_{\delta_k}),{B_r}\big)
\label{eq:policy-evolution}
\end{equation}
Unlike traditional post-training,
the goal of policy evolution is to improve compatibility between the reasoning policy and the execution harness instead of optimizing a universally stronger reasoner.

\paragraph{Overview.}

At a higher level,
each round alternates between
(i) evolving harness structures from trajectory-level execution evidence,
and
(ii) evolving harness-conditioned reasoning policies from the resulting survivor population:
\begin{equation}
\begin{aligned}
\mathbb{G}^{(r)}=
\{(\mathcal{H}_k^{(r)},\mathcal{R}^{(r)}_{\delta_k})\}
\rightarrow
 \{(\mathcal{H}_{k}^{(r+1)},\mathcal{R}^{(r)}_{\delta_k})\}\\
\rightarrow
\{(\mathcal{H}_{k}^{(r+1)},\mathcal{R}^{(r+1)}_{\delta_k})\}
=\mathbb{G}^{(r+1)}.
\label{eq:coevolution-process}
\end{aligned}
\end{equation}

By iterating this co-evolution process, both the execution harness and the governing reasoning policy evolve jointly, yielding increasingly adaptive agent systems over time. The implementation details of harness evolution and policy evolution are presented in
Sec.~\ref{sec:fault-guided-harness-tailoring} and Sec.~\ref{sec:trajectory-guided-reasoning-evolution},
respectively.

\subsection{Fault-Guided Harness Tailoring}
\label{sec:fault-guided-harness-tailoring}

The main challenge in agent system evolution is to locate which part of the harness causes failed execution. We therefore introduce a locate-and-refine mechanism to tailor better harnesses. We use \(T_{\psi}=(\mathbb{L}_{\omega},\mathbb{R}_{\omega},\Gamma_{\omega})\) to denote the overall evolution pipeline, where \(\mathbb{L}_{\omega}\) performs fault attribution, \(\mathbb{R}_{\omega}\) produces archive-guided improvement reports, \(\Gamma_{\omega}\) generates revised harness candidates.

\paragraph{Fault Attribution.}
For each active system \(\mathcal{G}_i^{(r)}=(\mathcal{H}_i^{(r)},\mathcal{R}^{(r)}_{\delta_i})\), HarnessForge first evaluates \(\mathcal{G}_i^{(r)}\) on batch \(B_r\), producing rollout traces \(\mathcal{T}_i^{(r)}\) and the evaluation vector \(\mathbf{J}(\mathcal{G}_i^{(r)};B_r)\) defined in Eq.~\ref{eq:batch-fitness}. A meta-agent then performs fault-attribution operation \(\mathbb{L}_{\omega}\) by jointly inspecting the current harness design and its representative failure trajectories:
\begin{equation}
    \mathbf{F}_{\mathcal{H}_i}^{(r)}
    =
    \mathbb{L}_{\omega}
    \left(
        \mathcal{H}_i^{(r)},
        \mathcal{T}_i^{(r)},
        \mathbf{J}(\mathcal{G}_i^{(r)};B_r)
    \right),
\label{eq:fault-attribution}
\end{equation}
where \(\mathbf{F}_{\mathcal{H}_i}^{(r)}\) is the fault report, attributing failures to planning, action and memory components.

\paragraph{Archive-Guided Improvement.}
HarnessForge maintains an archive \(\mathcal{Z}^{(r)}\) of historical harnesses, storing compact summaries of harness designs and their corresponding evaluation vectors \(\mathbf{J}\). Given the current harness \(\mathcal{H}_i^{(r)}\) and its fault report \(\mathbf{F}_{\mathcal{H}_i}^{(r)}\), the meta-agent samples exemplar cases \(\mathcal{S}_{\mathcal{H}_i}^{(r)}\subset\mathcal{Z}^{(r)}\) based on fault relevance and Pareto-front quality, and produces an improvement report:
\begin{equation}
    \mathbf{I}_{\mathcal{H}_i}^{(r)}
    =
    \mathbb{R}_{\omega}
    \left(
        \mathcal{H}_i^{(r)},
        \mathbf{F}_{\mathcal{H}_i}^{(r)},
        \mathcal{S}_{\mathcal{H}_i}^{(r)}
    \right).
\label{eq:archive-guided-refinement}
\end{equation}
The improvement report summarizes the likely improvement directions for the current harness, such as which component should be edited and which historical designs should be referenced. It serves as the input to the subsequent generation operator.

\paragraph{Refine-and-Filter.}
Based on the improvement report, HarnessForge generates revised harness candidates. For each active harness \(\mathcal{H}_i^{(r)}\), the generation operator \(\Gamma_{\omega}\) proposes \(K_{\mathrm{gen}}\) revised harnesses:
\begin{equation}
    \mathcal{C}_{\mathcal{H}_i}^{(r)}
    =
    \Gamma_{\omega}
    \left(
        \mathcal{H}_i^{(r)},
        \mathbf{I}_{\mathcal{H}_i}^{(r)}
    \right),
    ~~
    \left|\mathcal{C}_{\mathcal{H}_i}^{(r)}\right|
    =
    K_{\mathrm{gen}} .
\label{eq:candidate-generation}
\end{equation}
The operator only edits the execution-layer components \(\mathcal{P}\), \(\mathcal{A}\), and \(\mathcal{M}\). Let \(\mathcal{C}_0^{(r)}=\bigcup_i\mathcal{C}_{\mathcal{H}_i}^{(r)}\) be the pooled candidate set. Since fully evaluating every candidate is expensive, HarnessForge applies \textit{half-selection} over progressively larger task subsets. At filtering stage \(t\), each candidate harness \(\mathcal{H}\in\mathcal{C}_{t-1}^{(r)}\) is paired with the corresponding policy to form an executable system \(\mathcal{G}_{\mathcal{H}}\) and evaluated on \(B_{r,t}\) using the batch fitness \(\mathbf{J}(\mathcal{G}_{\mathcal{H}};B_{r,t})\). Candidates are selected according to Pareto optimality over the dimensions of this evaluation vector, and the retained subset becomes \(\mathcal{C}_t^{(r)}\).
After filtering, HarnessForge updates the archive \(\mathcal{Z}^{(r+1)}\) with each evaluated harness and its evaluation vector \(\mathbf{J}(\mathcal{G}_{\mathcal{H}};B_{r,t})\), enabling later Pareto-aware retrieval.
The survivor harnesses \(\mathcal{C}^{(r+1)}=\mathcal{C}_{T}^{(r)}\) are then passed to policy evolution. Implementation details of the whole tailoring process are provided in App.~\ref{app:meta-tailoring-operator}.


\begin{table*}[t]
\caption{Main results across benchmark groups for the Qwen3-4B and Qwen3-8B backbones. \legendbox{HFSearchBlue}{pale blue} rows indicate search-style methods, \legendbox{HFTrainBlue}{soft blue} rows indicate training-style methods, and \legendbox{HFForgePurple}{blue-purple} rows indicate our HarnessForge framework. Detailed benchmark and baseline configurations are provided in App.~\ref{app:benchmarks} and App.~\ref{app:baseline-config}.}
\centering
\footnotesize
\setlength{\tabcolsep}{1.9pt}
\renewcommand{\arraystretch}{1.05}
\begin{tabularx}{\textwidth}{@{}lYYYYYYYYYY@{}}
\toprule[1.15pt]
\multirow{2}{*}{\textbf{Method}} &
\multicolumn{2}{c}{\textbf{ToolHop}} &
\multicolumn{3}{c}{\textbf{SearchQA}} &
\multicolumn{2}{c}{\textbf{TMDB}} &
\multicolumn{3}{c}{\textbf{API-Bank}} \\
\cmidrule(lr){2-3}\cmidrule(lr){4-6}\cmidrule(lr){7-8}\cmidrule(lr){9-11}
& \textbf{Ans.} & \textbf{Path} & \textbf{Hotpot} & \textbf{2Wiki} & \textbf{Overall} & \textbf{Succ.} & \textbf{Path} & \textbf{Succ.} & \textbf{Path} & \textbf{API.} \\
\midrule[0.65pt]
\multicolumn{11}{c}{\textbf{Qwen3-4B}} \\
\midrule[0.4pt]
\rowcolor{HFSearchBlue} ADAS & 40.00 & 44.99 & 28.67 & 28.00 & 28.33 & 45.00 & 57.43 & 51.75 & 57.31 & 60.99 \\
\rowcolor{HFSearchBlue} AgentSquare & 29.23 & 39.46 & 29.33 & 30.00 & 29.67 & 35.00 & 49.83 & 39.47 & 47.82 & 46.10 \\
\rowcolor{HFSearchBlue} AFlow & 31.28 & 43.81 & 30.67 & 29.33 & 30.00 & 32.00 & 47.64 & 37.72 & 45.72 & 41.13 \\
\rowcolor{HFSearchBlue} MaAS & 42.05 & 51.80 & 30.67 & 30.00 & 30.33 & 37.00 & 46.57 & 52.63 & 62.74 & 63.83 \\
\rowcolor{HFSearchBlue} MermaidFlow & 44.62 & 58.74 & 31.33 & 28.00 & 29.67 & 39.00 & 51.95 & 54.39 & 64.33 & 60.28 \\
\addlinespace[1pt]
\rowcolor{HFTrainBlue} SFT & 45.13 & 63.90 & 36.67 & 38.67 & 37.67 & 61.00 & 75.70 & 69.30 & 75.10 & 73.76 \\
\rowcolor{HFTrainBlue} RLOO & 46.15 & 64.87  & 35.33 & 40.00 & 37.67 & 61.00 & 77.42 & 72.81 & 78.82 & 76.60 \\
\rowcolor{HFTrainBlue} GRPO & 49.74 & 66.41 & 36.00 & \textbf{43.33} & 39.67  & 64.00 & 76.67 & 71.93 & 76.46 & 75.18 \\
\addlinespace[1pt]
\rowcolor{HFForgePurple} HarnessForge & \textbf{52.82} & \textbf{68.10} & \textbf{42.00} & 42.00 & \textbf{42.00} & \textbf{76.00} & \textbf{85.10} & \textbf{77.19} & \textbf{80.07} & \textbf{82.27} \\
\midrule[0.65pt]
\multicolumn{11}{c}{\textbf{Qwen3-8B}} \\
\midrule[0.4pt]
\rowcolor{HFSearchBlue} ADAS & 42.05 & 61.32 & 31.33 & 32.00 & 31.67 & 51.00 & 60.41 & 54.39 & 60.14 & 59.57 \\
\rowcolor{HFSearchBlue} AgentSquare & 30.77 & 41.22 & 29.33 & 30.00 & 29.50 & 38.00 & 53.83 & 45.61 & 54.37 & 53.90 \\
\rowcolor{HFSearchBlue} AFlow & 32.82 & 54.12 & 31.33 & 32.67 & 32.00 & 39.00 & 51.28 & 43.86 & 49.74 & 47.52 \\
\rowcolor{HFSearchBlue} MaAS & 44.62 & 51.80 & 32.00 & 32.00 & 32.00 & 43.00 & 54.78 & 57.89 & 66.72 & 60.28 \\
\rowcolor{HFSearchBlue} MermaidFlow & 47.39 & 59.31 & 32.67 & 30.67 & 31.67 & 47.00 & 58.67 & 57.02 & 67.46 & 62.41 \\
\addlinespace[1pt]
\rowcolor{HFTrainBlue} SFT & 48.72 & 72.20 & 40.67 & 41.33 & 41.00 & 69.00 & 82.92 & 68.42 & 73.80 & 73.05 \\
\rowcolor{HFTrainBlue} RLOO & 50.77 & 72.75 & 40.00 & 39.33 & 39.67 & 74.00 & 84.25 & 66.67 & 73.76 & 71.63 \\
\rowcolor{HFTrainBlue} GRPO & 51.28 & \textbf{75.62} & 41.33 & 42.00 & 41.67 & 70.00 & 83.20  & 69.30 & 74.71 & 73.76 \\
\addlinespace[1pt]
\rowcolor{HFForgePurple} HarnessForge & \textbf{54.87} & 74.05 & \textbf{41.33} & \textbf{44.33} & \textbf{42.83} & \textbf{80.00} & \textbf{88.75}  & \textbf{74.56} & \textbf{78.22} & \textbf{78.01} \\
\bottomrule[1.15pt]
\end{tabularx}
\vskip -0.1in
\label{tab:main-result}
\end{table*}


\subsection{Harness-Conditioned Policy Alignment}
\label{sec:trajectory-guided-reasoning-evolution}
After harness evolution, each survivor harness must be paired with an executor that can reliably operate under its evolved planning, action, and memory interface. We relabel the survivor set \(\mathcal{C}^{(r+1)}=\{\mathcal{H}_k^{(r+1)}\}_k\) by lineage, so each survivor inherits the corresponding round-\(r\) parent policy \(\mathcal{R}^{(r)}_{\delta_k}\). Its goal is not to train a universally stronger reasoner, but to align the inherited policy with the execution conventions induced by a particular harness.

\paragraph{Parent Initialization.}
For each survivor harness \(\mathcal{H}_k^{(r+1)}\), HarnessForge initializes the child policy from its parent lineage and trains a new harness-specific LoRA~\citep{hu2021loralowrankadaptationlarge} update, denoted as \(\delta_k^{(r+1)}=\mathrm{Merge}(\delta_k^{(r)})\oplus\Delta\delta_k^{(r+1)}\). Here, \(\mathrm{Merge}(\cdot)\) folds the parent adapter into the child initialization, while \(\Delta\delta_k^{(r+1)}\) adapts the policy to the evolved harness. Details are provided in App.~\ref{app:policy-lineage}.

\paragraph{Trajectory Curation.} 
Under the same rollout budget, policy evolution should not introduce a separate data-collection stage. HarnessForge therefore reuses the rollout pool already produced when \(\mathcal{H}_k^{(r+1)}\) is evaluated during budgeted harness selection. Denote this pool by \(\mathcal{T}_k^{(r+1)}\). We keep the successful trajectories:
\begin{equation}
    \mathcal{T}_{k}^{+}
    =
    \{\tau_x\in\mathcal{T}_k^{(r+1)} \mid S(\tau_x)=1\}.
\label{eq:trajectory-filter}
\end{equation}
Here \(S(\tau_x)\) is the task-success indicator used in evaluation. This keeps trajectory curation tied to the same success signal that determines the harness success rate, while avoiding extra rollout cost.

\paragraph{Harness-Conditioned Evolution.}
HarnessForge converts the retained successful rollouts into step-level supervision for the selected harness. Each trajectory \(\tau_x\in\mathcal{T}_k^{+}\) is decomposed into decision pairs \((z_t,y_t)\) over its time steps, yielding the harness-conditioned dataset \(\mathcal{D}_{\mathcal{H}_k}=\{(z_t,y_t)\}\) used by the alignment loss. The input \(z_t=\bigl(x,\mathcal{H}_k^{(r+1)},o_{\leq t},m_t,a_t\bigr)\) packages the task instruction, active harness interface, accumulated observations, current memory state, and available actions; the target \(y_t\) is the corresponding next behavior, such as a reasoning step, tool action, memory operation, or final response. More general policy-update objectives can instantiate this stage, as discussed in App.~\ref{app:policy-alignment-instantiations}. In the main experiments, HarnessForge uses supervised trace alignment because it reuses the successful trajectories above and offers a favorable rollout--performance tradeoff::
\begin{equation}
    \Delta\delta_k^{(r+1)}
    =
    \arg\min_{\Delta\delta}
    \mathcal{L}\big(\mathcal{R}^{(r)}_{\delta_k}\oplus\Delta\delta;\mathcal{D}_{\mathcal{H}_k}\big).
\label{eq:sft-loss}
\end{equation}
This objective learns only an incremental adapter on top of the inherited policy, aligning the executor to \(\mathcal{H}_k^{(r+1)}\) without spending additional rollouts. The result is a matched next-round harness--policy pair \(\mathcal{G}_k^{(r+1)}=(\mathcal{H}_k^{(r+1)},\mathcal{R}^{(r+1)}_{\delta_k})\), which enters the next-round population rather than being treated as a general-purpose model upgrade.

\section{Experiments}
\label{sec:experiments}

\subsection{Experimental Setup}

\paragraph{Benchmarks.}
We evaluate HarnessForge on diverse benchmarks: ToolHop~\citep{ye-etal-2025-toolhop}, RestBench-TMDB~\citep{song2023restgptconnectinglargelanguage}, and API-Bank~\citep{li-etal-2023-api} measure tool selection, and API-grounded execution capabilities. SearchQA, built from HotpotQA~\citep{yang-etal-2018-hotpotqa} and 2WikiMultiHopQA~\citep{ho-etal-2020-constructing}, evaluates retrieval-heavy multi-hop question answering. Detailed descriptions and dataset statistics are in App.~\ref{app:benchmarks}.
\vspace{-1mm}
\paragraph{Implementation Details.}
We instantiate HarnessForge with Qwen3-4B and Qwen3-8B~\citep{yang2025qwen3technicalreport} as the default backbones to evaluate its effectiveness. The meta-agent used for harness evolution is GPT-5.5. We run $r=3$ evolution rounds and report the main setting that retains $|C|=2$ survivor harnesses per round. Additional implementation details are provided in App.~\ref{app:method-config}.
\vspace{-1mm}
\paragraph{Baselines.}
We compare HarnessForge against two main baseline groups: Search-style harness baselines include AFlow~\citep{zhang2025aflow}, ADAS~\citep{ICLR2025_36b7acf6}, AgentSquare~\citep{shang2025agentsquare}, MaAS~\citep{zhang2025multiagent}, and MermaidFlow~\citep{zheng2025mermaidflowredefiningagenticworkflow}. Training-style baselines include SFT, GRPO~\citep{shao2024deepseekmathpushinglimitsmathematical}, and RLOO~\citep{ahmadian2024basicsrevisitingreinforcestyle}. Detailed configurations of baselines are provided in App.~\ref{app:baseline-config}.

\begin{figure*}[!t]
  \centering
  \begin{subfigure}[t]{0.495\textwidth}
    \centering
    \includegraphics[width=\linewidth]{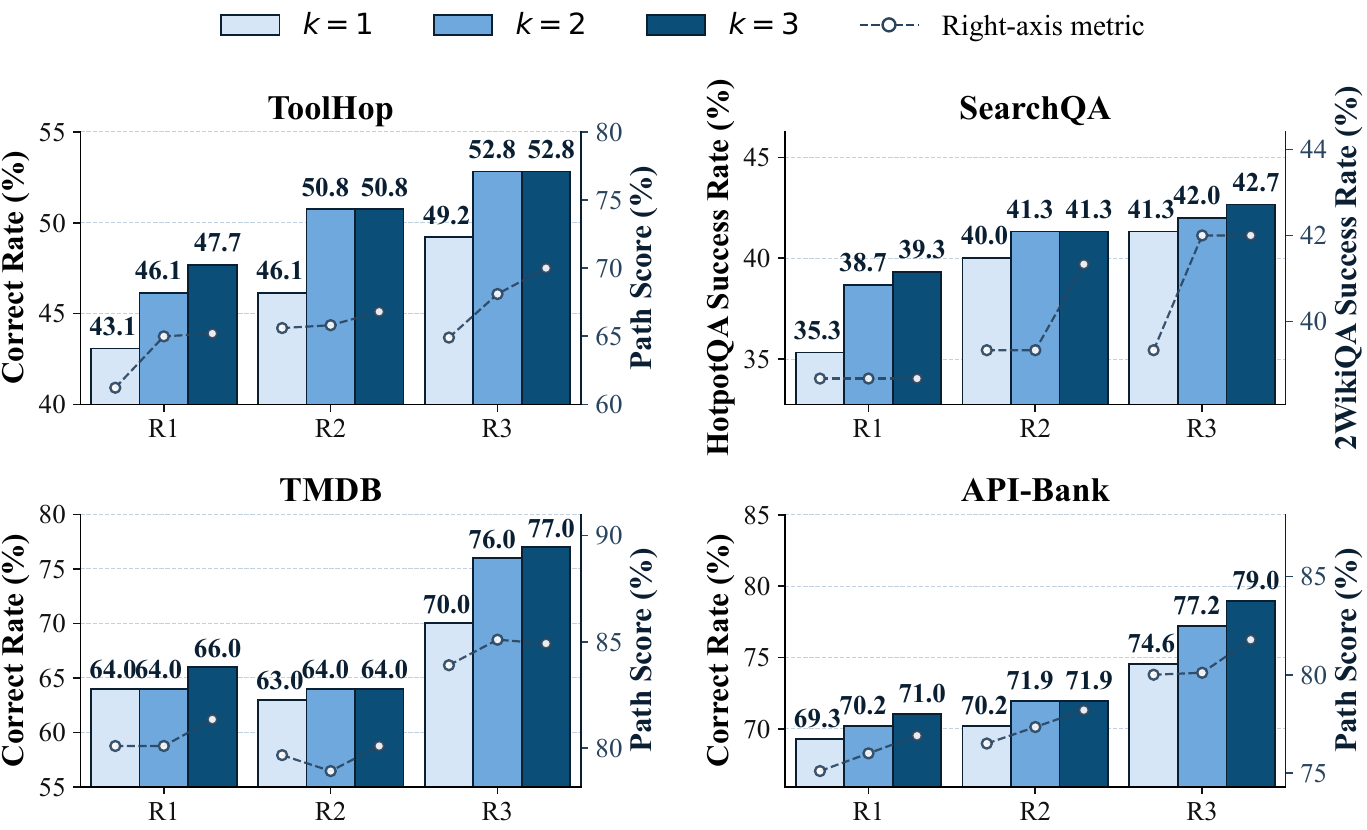}
    \caption{Retained-harness sensitivity across four benchmark groups.}
    \label{fig:sensitivity-retained-k-4bench}
  \end{subfigure}
  \hfill
  \begin{subfigure}[t]{0.495\textwidth}
    \centering
    \includegraphics[width=\linewidth]{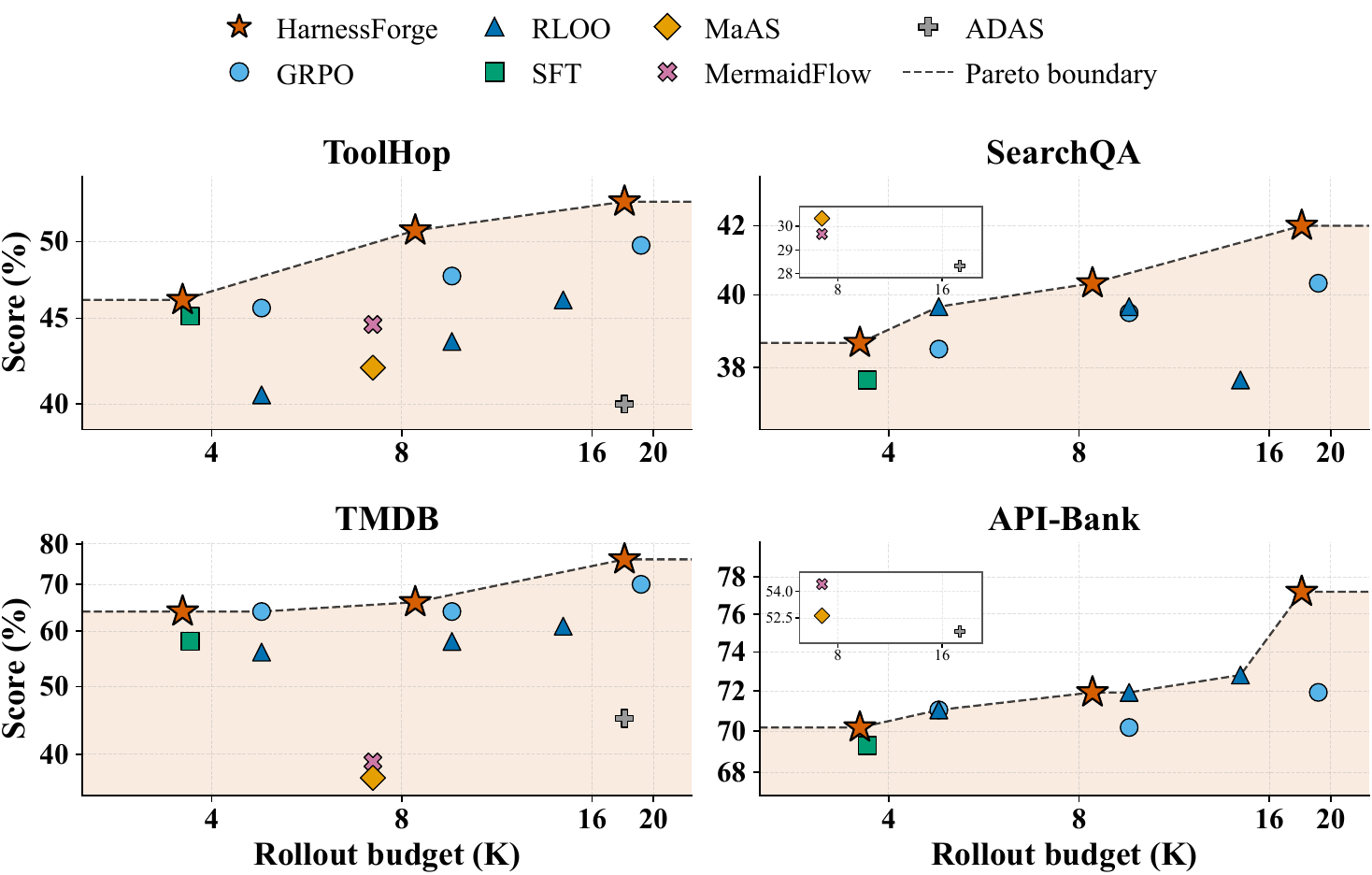}
    \caption{Budget--performance Pareto analysis across benchmarks.}
    \label{fig:budget-pareto-4bench}
  \end{subfigure}
  \caption{Framework analysis of HarnessForge. \textbf{Left:} performance sensitivity to the number of retained harnesses per evolution round. \textbf{Right:} rollout-budget efficiency compared with alternative adaptation settings.}
  \label{fig:framework-analysis-tradeoffs}
  \vskip -0.2in
\end{figure*}

\subsection{Main Results}

Tab.~\ref{tab:main-result} compares HarnessForge with harness-search and policy-training baselines across five agentic benchmark and two backbone sizes, averaging +3.56\% over per-metric strongest baselines.
HarnessForge delivers strong performance gains and reaches SOTA results on most benchmarks, spanning both tool-use and retrieval settings. Notably, the policy-training baselines RLOO and GRPO require larger rollout budgets than HarnessForge (App.~\ref{app:training-style-baselines}), yet still fall behind on most metrics.
The largest gains appear on TMDB: HarnessForge improves success by 12.00\% with Qwen3-4B and 6.00\% with Qwen3-8B over the strongest baseline. On API-Bank, it improves API accuracy by average 4.96\% across backbones.
It also remains strong on reasoning-heavy benchmarks, with average 3.34\% ToolHop answer gains across backbones and best SearchQA overall scores 42.83\%.

\begin{table}[t]
\caption{Module ablation of HarnessForge using Qwen3-4B.
ToolHop reports final-answer correctness, and SearchQA reports the macro-average answer F1.}
\centering
\footnotesize
\setlength{\tabcolsep}{2.2pt}
\renewcommand{\arraystretch}{1.06}
\begin{tabular*}{\columnwidth}{@{\extracolsep{\fill}}c p{0.34\columnwidth}cc@{}}
\toprule[1.15pt]
\textbf{Round} &
\textbf{Variant} &
\multicolumn{1}{c}{\textbf{ToolHop}} &
\multicolumn{1}{c}{\textbf{SearchQA}} \\
\midrule[0.65pt]
Round0 & Vanilla & 41.03 & 34.33 \\
\midrule[0.65pt]
\multirow{3}{*}{Round1}
& HarnessForge & 46.15 & 38.67 \\
& \hspace{0.6em}\textit{-w/o Harness Evo.} & 43.08\drop{3.07} & 35.67\drop{3.00} \\
& \hspace{0.6em}\textit{-w/o Policy Evo.} & 44.62\drop{1.53} & 36.33\drop{2.34} \\
\midrule[0.35pt]
\multirow{3}{*}{Round2}
& HarnessForge & 50.77 & 40.33 \\
& \hspace{0.6em}\textit{-w/o Harness Evo.} & 44.62\drop{6.15} & 36.33\drop{4.00} \\
& \hspace{0.6em}\textit{-w/o Policy Evo.} & 48.72\drop{2.05} & 38.33\drop{2.00} \\
\midrule[0.35pt]
\multirow{3}{*}{Round3}
& HarnessForge & 52.82 & 42.00 \\
& \hspace{0.6em}\textit{-w/o Harness Evo.} & 46.67\drop{6.15} & 37.00\drop{5.00} \\
& \hspace{0.6em}\textit{-w/o Policy Evo.} & 50.26\drop{2.56} & 39.00\drop{3.00} \\
\bottomrule[1.15pt]
\end{tabular*}
\vskip -0.2in
\label{tab:ablation-results}
\end{table}

\subsection{Ablation Study}
Tab.~\ref{tab:ablation-results} shows that removing either harness tailoring or policy alignment consistently degrades performance on both ToolHop and SearchQA, indicating that both modules are necessary for HarnessForge. Harness tailoring is the dominant factor: disabling it causes the largest drop in every round, and the gap becomes larger as evolution proceeds, increasing from -3.07\%/-3.00\% in Round 1 to -6.15\%/-5.00\% in Round 3 on ToolHop/SearchQA. Removing policy alignment also hurts performance, with final-round drops of -2.56\% on ToolHop and -3.00\% on SearchQA, suggesting that the reasoner must adapt to the evolved execution interface. Overall, the widening gaps show that HarnessForge's gains come from harness-policy co-evolution.

\subsection{Framework Analysis}
\label{sec:framework-analysis}

\paragraph{Retained-harness sensitivity.}
Fig.~\ref{fig:sensitivity-retained-k-4bench} studies how the survivor pool size affects harness evolution. Retaining a single harness is often too restrictive: at the final round, increasing from \(k=1\) to \(k=2\) improves the main metric by \(3.6\%\) on ToolHop, \(0.7\%\) on SearchQA, \(6.0\%\) on TMDB, and \(2.6\%\) on API-Bank, averaging \(3.2\%\) points. Further increasing to \(k=3\) brings only marginal additional gains on most benchmarks, suggesting that excessive retention weakens selection pressure. Across rounds, \(k=2\) also yields consistent improvements from R1 to R3 on the four benchmarks groups. These results indicate that a small survivor population preserves useful harness diversity while still maintaining effective selection, enabling agent systems to emerge through population-level harness--policy evolution rather than one-shot scaffold optimization.

\paragraph{Rollout-budget efficiency.}
Fig.~\ref{fig:budget-pareto-4bench} compares different methods under varying rollout budgets. HarnessForge consistently lies on or near the Pareto frontier across all four benchmark groups, indicating that the proposed co-evolution process is both performance-effective and rollout-efficient. Unlike policy-only training methods that must spend additional rollouts to explore better behaviors under a fixed interface, HarnessForge reuses rollout evidence to improve the external harness and then aligns the policy to the selected interface. This allows useful structural revisions and harness-conditioned execution patterns to accumulate across rounds, reducing the need for large-scale exploration. Consequently, HarnessForge provides a stronger performance--budget tradeoff than other on-policy training-style baselines.

\begin{figure}[!t]
  \begin{center}
    \centerline{\includegraphics[width=0.98\columnwidth]{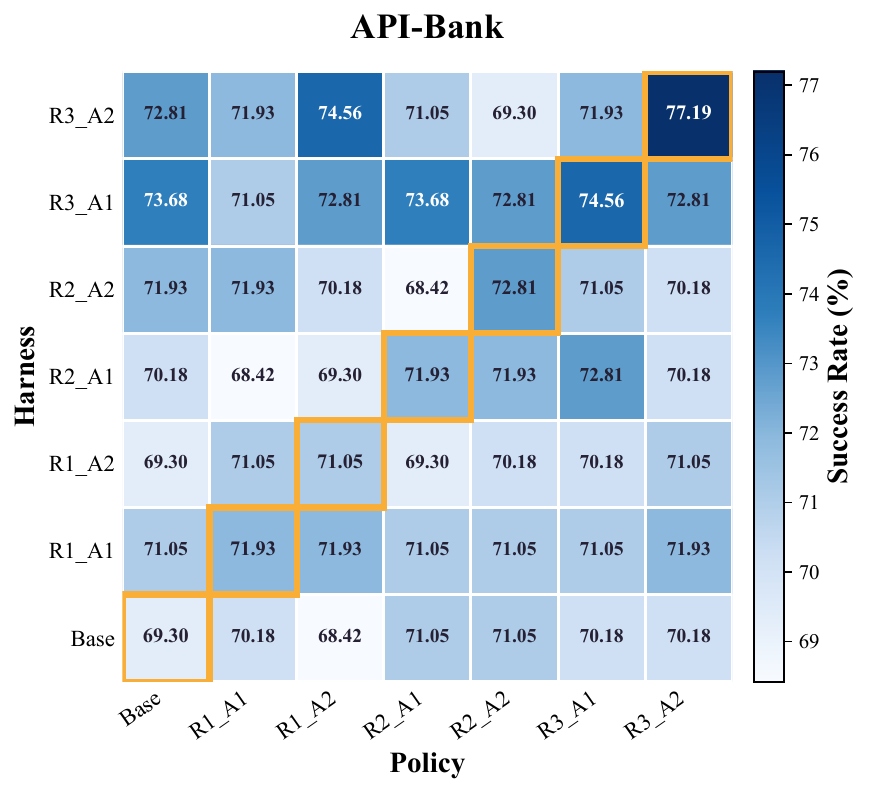}}
    \vskip -0.1in
    \caption{Harness--policy compatibility matrices on API-Bank. Rows denote evolved harnesses and columns denote evolved policy across evolution rounds.}
    \label{fig:compatibility_matrix_api-bank}
  \end{center}
  \vskip -0.4in
\end{figure}

\begin{figure*}[ht]
  \begin{center}
    \centerline{\includegraphics[width=2.1\columnwidth]{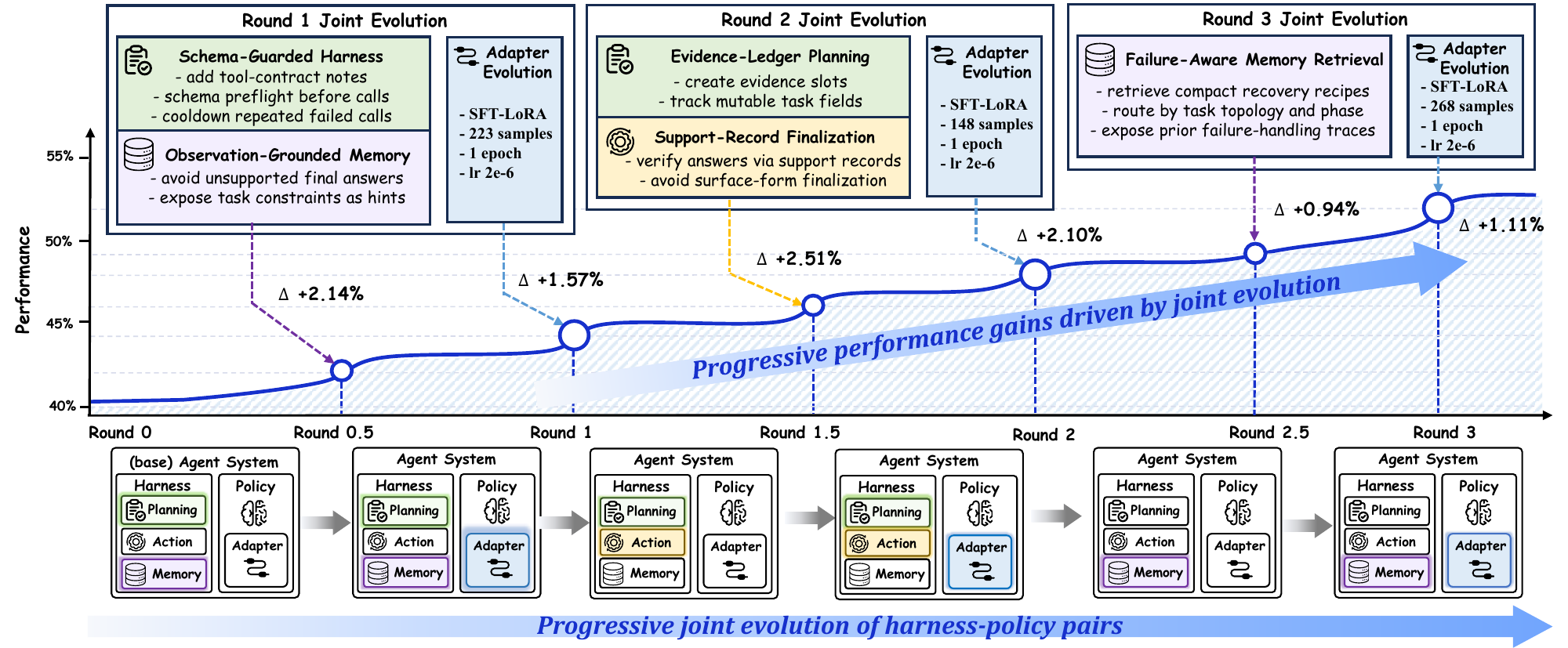}}
    \caption{Representative ToolHop lineage of HarnessForge across three harness--policy co-evolution rounds.}
    \label{fig:case_study}
  \end{center}
  \vskip -0.4 in
\end{figure*}

\paragraph{Adaptation necessity analysis.}
Fig.~\ref{fig:compatibility_matrix_api-bank} evaluates all cross-combinations of evolved harnesses and policy adapters on API-Bank. Along the matched diagonal, performance improves from \(69.30\%\) for the base pair to \(77.19\%\) for the final pair. Moreover, the final harness paired with earlier policies averages only \(71.93\%\), and the final policy paired with earlier harnesses averages only \(71.06\%\). These gaps indicate that HarnessForge does not simply produce independently stronger components, but instead reveals pair-specific compatibility gains. It induces harness-conditioned policy specialization, making matched harness--policy pairs substantially more effective than mismatched combinations. Additional compatibility matrices and detailed analysis are provided in App.~\ref{app:compatibility_matrix}.

\begin{table}[t]
\caption{Training agnostic analysis. We evaluate HarnessForge+SFT, HarnessForge+GRPO, and HarnessForge+RLOO on ToolHop, and report API-Bank as an auxiliary transfer/evaluation setting.}
\centering
\footnotesize
\setlength{\tabcolsep}{2.3pt}
\renewcommand{\arraystretch}{1.06}
\begin{tabular*}{\columnwidth}{@{\extracolsep{\fill}}clcccccc@{}}
\toprule[1.15pt]
\multirow{2}{*}{\textbf{Round}} &
\multirow{2}{*}{\textbf{Training}} &
\multicolumn{2}{c}{\textbf{ToolHop}} &
\multicolumn{3}{c}{\textbf{API-Bank}} &
\multirow{2}{*}{\makecell{\textbf{Budget}\\\textbf{(K)}}} \\
\cmidrule(lr){3-4}\cmidrule(lr){5-7}
& & \textbf{Ans.} & \textbf{Path} & \textbf{Succ.} & \textbf{Path} & \textbf{API.} & \\
\midrule[0.65pt]
\multirow{3}{*}{1} 
& SFT  & 45.13 & 60.66 & 69.30 & 74.85 & 73.05 & 2.40 \\
& GRPO & 45.64 & 59.49 & 70.18 & \textbf{76.02} & \textbf{74.47} & 7.20 \\
& RLOO & \textbf{46.67} & \textbf{61.17} & \textbf{70.18} & 75.73 & 73.76 & 7.20 \\
\midrule[0.35pt]
\multirow{3}{*}{2} 
& SFT  & 48.72 & 63.70 & 71.05 & 75.15 & \textbf{75.89} & 5.60 \\
& GRPO & 49.23 & \textbf{64.28} & 69.30 & 76.90 & 73.76 & 20.00 \\
& RLOO & \textbf{49.74} & 63.80 & \textbf{71.05} & \textbf{77.78} & 75.18 & 20.00 \\
\midrule[0.35pt]
\multirow{3}{*}{3} 
& SFT  & 50.77 & 65.87 & 71.05 & 76.90 & 75.18 & 12.00 \\
& GRPO & \textbf{52.31} & \textbf{67.32} & 71.93 & 77.78 & 75.18 & 45.60 \\
& RLOO & 51.28 & 66.45 & \textbf{72.80} & \textbf{79.09} & \textbf{76.60} & 45.60 \\
\bottomrule[1.15pt]
\end{tabular*}
\label{tab:training-method-agnosticism}
\vskip -0.2in
\end{table}

\vspace{-1mm}
\paragraph{Training-method agnosticism.}
HarnessForge is not tied to a specific policy-training method. In the main experiments, we use supervised fine-tuning (SFT) as the default instantiation because it directly internalizes harness-induced execution patterns from curated trajectories. Tab.~\ref{tab:training-method-agnosticism} compares SFT, GRPO, and RLOO within the same harness--policy co-evolution framework. Across rounds, replacing SFT with GRPO or RLOO further improves several metrics, e.g., in Round 3 GRPO improves ToolHop answer accuracy from 50.77\% to 52.31\%, while RLOO improves API-Bank success from 71.05\% to 72.80\%. These gains, however, require substantially larger rollout budgets: Round-3 RL-style instantiations use 45.6K rollouts compared with 12.0K for SFT. This suggests that HarnessForge's framework-level gains are not specific to SFT, while SFT remains a rollout-efficient default and RL-style objectives provide additional improvement potential at higher cost.

\subsection{Case Study}
Fig.~\ref{fig:case_study} visualizes a representative ToolHop lineage of HarnessForge across three harness--policy co-evolution rounds. Starting from the base agent, Round 1 improves task decomposition and memory exposure by introducing finer-grained subgoals, and more consistent context injection, yielding a \(2.14\%\) performance gain and the subsequent adapter alignment further contributes \(1.57\%\). Round 2 focuses on planning and action reliability, followed by a matched adapter update that together adds \(2.51\%\) and \(2.10\%\). Round 3 targets memory retrieval, improving retrieval relevance, reducing noisy memory access with the final adapter update adding another \(0.94\%\) and \(1.11\%\). This lineage shows that HarnessForge's performance improves progressively as the harness is tailored and the policy is aligned to the evolving execution interface. Additional case studies are provided in App.~\ref{app:case_study}.

\section{Conclusion}
We introduced HarnessForge, a meta-adaptive framework that reformulates LLM agent adaptation as harness--policy pair evolution. Instead of optimizing external workflows or internal policies in isolation, HarnessForge co-evolves the execution harness and the reasoning policy through fault-guided harness tailoring and harness-conditioned policy alignment. Experiments across diverse agent benchmarks show consistent gains over diverse baselines, favorable rollout--performance tradeoffs, and strong matched-pair compatibility. These results demonstrate that effective agent-system adaptation depends on optimizing the executable compatibility between the harness and the policy.

\section*{Limitations}

HarnessForge is primarily evaluated with Qwen3-4B and Qwen3-8B backbones. This setting is important for resource-constrained agent deployment, where the coupling between the external harness and the internal policy is especially salient: the harness can provide structural support for limited model capabilities, while the policy must learn to execute the harness-induced paradigm. Whether the same magnitude of harness--policy compatibility gains holds for substantially larger frontier-scale models remains an open direction.

HarnessForge also requires repeated rollouts for harness profiling, selection, and policy alignment. Although our design reuses rollout trajectories across harness evolution and policy alignment to improve rollout efficiency, long-horizon environments can still make the evolution process costly. Future work could reduce this cost through proxy evaluation, adaptive rollout allocation, or learned early-stopping criteria.

Finally, the HarnessForge currently uses a structured meta-evolution protocol with constrained edit operators over planning, action, and memory components. This design improves executability and makes the evolution process auditable, but it does not exhaustively explore the full space of possible agent-system implementations, such as arbitrary code-level harness rewrites, new tool abstractions, or learned verifier modules. Future work could extend the operator space and study cost-aware meta-evolution with alternative closed- or open-source meta-agents.

\bibliography{custom}

\newpage

\appendix



\section{Algorithm and Notation}
\label{app:algorithm-notation}

\begin{algorithm}[h]
\caption{HarnessForge Co-evolution Round}
\label{alg:harnessforge}
\begin{algorithmic}[1]
\Require parent population $\mathcal{G}^{(r)}$, evolution batch $B_r$, archive $\mathcal{Z}^{(r)}$
\For{each parent pair $G_i^{(r)}=(H_i^{(r)},R_{\delta_i}^{(r)})$}
    \State Roll out $G_i^{(r)}$ on $B_r$ and collect traces $\mathcal{T}_i^{(r)}$
    \State Generate fault report $F_{H_i}^{(r)}$
    \State Retrieve archive cases and produce improvement brief $\mathbf{I}_{H_i}^{(r)}$
    \State Generate $K_{\mathrm{gen}}$ child harnesses
    \State Discard invalid children using interface and smoke tests
\EndFor
\State Select survivor harnesses by staged Pareto filtering over $B_{r,1},\ldots,B_{r,T}$
\For{each survivor harness $H_k^{(r+1)}$}
    \State Reuse successful rollout traces from filtering to construct $D_{H_k}$
    \State Materialize an independent copy of the parent lineage policy and train a new harness-specific adapter
    \State Return matched pair $(H_k^{(r+1)},R_{\delta_k}^{(r+1)})$
\EndFor
\State Update archive with evaluated harnesses, metrics, reports, and selection logs
\end{algorithmic}
\end{algorithm}

\subsection{Notation}

Tab.~\ref{tab:notation} summarizes the notation used in the main text and appendix.

\begin{table}[h]
\caption{Notation used by HarnessForge.}
\centering
\small
\begin{tabular}{@{}p{0.24\linewidth}p{0.66\linewidth}@{}}
\hline
\textbf{Symbol} & \textbf{Meaning} \\
\hline
\(\mathcal{G}=(\mathcal{H},\mathcal{R}_{\delta})\) & Agent system consisting of an executable harness and an adapted reasoner. \\
\(\mathcal{H}=(\mathcal{P},\mathcal{A},\mathcal{M})\) & Harness with planning, action, and memory components. \\
\(\mathcal{R}_{\theta_0}\) & Frozen base reasoner before harness-conditioned adaptation. \\
\(\delta\) & Lightweight adapter parameters associated with a policy lineage. \\
\(\mathbb{G}^{(r)}\) & Population of harness--policy pairs at evolution round \(r\). \\
\(B_r, B_{r,t}\) & Evolution batch at round \(r\), and the subset used at filtering stage \(t\). \\
\(\mathcal{T}_i^{(r)}\) & Rollout traces collected by pair \(i\) at round \(r\). \\
\(\mathbf{J}(\mathcal{G};B)\) & Multi-objective evaluation vector over task performance and efficiency. \\
\(\mathcal{Z}^{(r)}\) & Archive of evaluated harnesses, rollout summaries, diagnostics, and selection records. \\
\(\mathbf{F}_{\mathcal{H}}^{(r)}\), \(\mathbf{I}_{\mathcal{H}}^{(r)}\) & Fault report and archive-guided improvement brief for harness \(\mathcal{H}\). \\
\(\mathcal{C}^{(r+1)}\) & Survivor harness set passed to policy alignment for the next round. \\
\hline
\end{tabular}
\label{tab:notation}
\end{table}

\section{Datasets Details}
\label{app:benchmarks}
\subsection{Evaluation Benchmark and Metrics}
The main experiments cover five datasets organized into four benchmark families. ToolHop, RestBench-TMDB, and API-Bank each form one benchmark family. SearchQA is the retrieval-heavy benchmark family and contains two datasets, HotpotQA and 2WikiMultiHopQA. We therefore use ``five datasets'' when describing dataset coverage and ``four benchmark families'' or ``four benchmark groups'' when referring to the columns in Tab.~\ref{tab:main-result}.

Let \(\mathcal{D}\) denote the evaluated split, \(N=|\mathcal{D}|\), \(\hat{a}_i\) and \(a_i\) denote the predicted and gold final answers for instance \(i\), and \(\hat{\pi}_i\) and \(\pi_i\) denote the predicted and gold tool/API-call paths. We use \(\mathbf{1}[\cdot]\) for the indicator function. For answer matching, \(\operatorname{norm}(\cdot)\) lowercases text and removes articles, punctuation, and extra whitespace, following standard QA evaluation.

\paragraph{ToolHop.}
ToolHop evaluates multi-hop tool use, where each instance requires the agent to decompose a question, call tools for intermediate evidence, and return a final answer~\citep{ye-etal-2025-toolhop}. Our current held-out split contains 195 evaluated instances. We report \textbf{Correct}, the final-answer correctness judged by the benchmark evaluator. Let \(m_i^{\mathrm{ans}}=1\) iff the normalized prediction matches the gold answer. Then
\begin{equation}
    \mathrm{Correct}_{\mathrm{ToolHop}}
    = \frac{\sum_{i=1}^{N}
    m_i^{\mathrm{ans}}}{N},
\end{equation}
and \textbf{Path}, the average fraction of required intermediate subgoals solved during the trajectory. If \(G_i\) is the set of required intermediate subgoals and \(\hat{G}_i\) is the set credited by the evaluator, then
\begin{equation}
    \mathrm{Path}_{\mathrm{ToolHop}}
    = \frac{1}{N}\sum_{i=1}^{N}
    \frac{|\hat{G}_i\cap G_i|}{|G_i|}.
\end{equation}

\paragraph{SearchQA.}
SearchQA evaluates retrieval-heavy multi-hop question answering over local evidence corpora constructed from HotpotQA and 2WikiMultiHopQA~\citep{yang-etal-2018-hotpotqa,ho-etal-2020-constructing}. We use the normalized token-level answer F1 score. For token multisets \(T(\hat{a}_i)\) and \(T(a_i)\), let
\begin{equation}
\begin{aligned}
    P_i &= \frac{|T(\hat{a}_i)\cap T(a_i)|}{|T(\hat{a}_i)|}, \\
    R_i &= \frac{|T(\hat{a}_i)\cap T(a_i)|}{|T(a_i)|}, \\
    F1_i &= \frac{2P_iR_i}{P_i+R_i}.
\end{aligned}
\end{equation}
The reported HotpotQA and 2WikiMultiHopQA scores are
\begin{equation}
\begin{aligned}
    \mathrm{Score}_{d}
    &=\frac{\sum_{i\in\mathcal{D}_{d}} F1_i}{|\mathcal{D}_{d}|}, \\
    d&\in\{\mathrm{Hotpot},\mathrm{2Wiki}\},
\end{aligned}
\end{equation}
and \textbf{Overall} is the macro-average over the two subsets:
\begin{equation}
    \mathrm{Overall}_{\mathrm{SearchQA} }
    = \frac{\mathrm{Score}_{\mathrm{Hotpot}}+\mathrm{Score}_{\mathrm{2Wiki}}}{2}.
\end{equation}

\paragraph{RestBench-TMDB.}
RestBench-TMDB evaluates REST-style API use over a movie-database environment~\citep{song2023restgptconnectinglargelanguage}. Each task specifies an information need that must be satisfied through one or more API calls. Our current evaluation split contains 100 instances. We report \textbf{Success}, the fraction of tasks whose final answer satisfies the verifier,
\begin{equation}
    \mathrm{Success}_{\mathrm{TMDB}}
    = \frac{\sum_{i=1}^{N} v_i}{N},
    \quad v_i\in\{0,1\},
\end{equation}
and \textbf{Path}, the evaluator's path-rate metric for matching the required API execution path. Following the RestBench notion of a correct API path, we count a path as matched if the gold API-call sequence is preserved as an ordered subsequence of the predicted path:
\begin{equation}
    \mathrm{Path}_{\mathrm{TMDB}}
    = \frac{1}{N}\sum_{i=1}^{N}
    \mathbf{1}\!\left[\pi_i\preceq\hat{\pi}_i\right],
\end{equation}
where \(\preceq\) denotes ordered-subsequence matching.

\paragraph{API-Bank.}
API-Bank evaluates structured API calling across diverse user requests~\citep{li-etal-2023-api}. Each instance provides a user instruction and a gold API-call specification, including the function name and arguments. Our current evaluation split contains 114 instances. We report \textbf{Success}, the fraction of instances whose full call trajectory and final response satisfy the evaluator,
\begin{equation}
    \mathrm{Success}_{\mathrm{APIBank}}
    = \frac{1}{N}\sum_{i=1}^{N}
    \mathbf{1}\!\left[\operatorname{Eval}(\hat{\pi}_i,\hat{a}_i)=1\right].
\end{equation}
We report \textbf{Path} as an ordered API-name overlap score. Let \(L_i\) be the length of the longest common subsequence between the predicted and gold API-name sequences, \(P_i^{\pi}=L_i/|\hat{\pi}_i|\), and \(R_i^{\pi}=L_i/|\pi_i|\). Then
\begin{equation}
    \mathrm{Path}_{\mathrm{APIBank}}
    = \frac{1}{N}\sum_{i=1}^{N}
    \frac{2P_i^{\pi}R_i^{\pi}}{P_i^{\pi}+R_i^{\pi}}.
\end{equation}
Finally, \textbf{API Accuracy} measures call-level exactness after aligning predicted calls to gold calls. Let \(m_{i,t}^{\mathrm{api}}=1\) iff the aligned predicted call has the correct API name and schema-normalized arguments, i.e., \(\operatorname{name}(\hat{c}_{i,t})=\operatorname{name}(c_{i,t})\) and \(\operatorname{args}(\hat{c}_{i,t})\simeq\operatorname{args}(c_{i,t})\). Then
\begin{equation}
    \mathrm{APIAcc}_{\mathrm{APIBank}}
    = \frac{1}{\sum_i |\pi_i|}\sum_{i=1}^{N}\sum_{t=1}^{|\pi_i|}
    m_{i,t}^{\mathrm{api}},
\end{equation}
where \(\simeq\) denotes schema-normalized argument equivalence.

\subsection{Training \& Evolution Data}
We construct a 3.8K(3,800)-task training pool covering three complementary agent capabilities: general tool use, complex environment interaction, and offline retrieval.
The training pool is strictly disjoint from all held-out evaluation splits used in the main experiments. Training tasks are used only for HarnessForge harness evolution, trajectory curation, and policy-adapter training, as well as for training the training-style baselines (SFT, GRPO, and RLOO); the test splits are reserved for final evaluation.

\begin{itemize}[leftmargin=1.35em, labelsep=0.45em, itemsep=1pt, topsep=2pt, parsep=0pt]
    \item \textbf{Complex Environment Interaction:} We sample 2.0k tasks from EnvScaler-RL~\citep{song2026envscalerscalingtoolinteractiveenvironments}. Each task provides an executable environment, an initial configuration, a task instruction, available tools, and verifier-based feedback. This subset targets multi-step tool use in stateful environments.

    \item \textbf{General Tool-Use:} We sample 0.8k instances from ToolHop~\citep{ye-etal-2025-toolhop}. These tasks emphasize tool selection, tool chaining, and precise function-call execution.

    \item \textbf{Offline QA \& Retrieval:} We sample 1.0k tasks from Wikipedia-based QA datasets, including Natural Questions (NQ)~\citep{kwiatkowski2019natural}, HotpotQA~\citep{yang-etal-2018-hotpotqa}, and 2WikiMultiHopQA~\citep{ho-etal-2020-constructing}. In our setup, we convert these datasets into an offline/local-corpus retrieval setting: each question is paired with a fixed local document collection, and the model must retrieve supporting evidence from that corpus before answering. This subset trains reproducible evidence retrieval and answer generation without relying on live web access.
\end{itemize}

Overall, the resulting training pool contains \textbf{3.8k samples} and provides a compact mixture of tool-use, environment interaction, and offline retrieval tasks.

\subsection{Split and Deduplication Protocol}

All reported test scores are computed on held-out evaluation splits. The 3.8K training and evolution pool is used for harness evolution, trajectory curation, and policy-adapter training, but is disjoint from the held-out test splits used for final reporting. When a source dataset contributes to both training/evolution and evaluation, we use non-overlapping split identifiers and remove potential duplicates by dataset identifier and normalized task instruction. For converted offline retrieval tasks, the local evidence corpus is fixed before adaptation and the held-out question identifiers are not exposed during harness search, policy training, or model selection.

\section{Harness Tailoring Details}
\label{app:method-config}

\subsection{Evaluator Setting}
\label{app:Evaluator}

HarnessForge uses a multi-objective evaluator to select harness candidates and
retrieve useful historical cases from the archive. We reuse the benchmark metrics
defined in App.~\ref{app:benchmarks} as the task-performance dimension, and add
efficiency dimensions for candidate selection. For each candidate agent system
\(\mathcal{G}\) and evaluation batch \(B\), we compute
\begin{equation}
\begin{aligned}
\mathbf{J}(\mathcal{G};B)
=
\big(&
\operatorname{Perf}(\mathcal{G};B),
-\operatorname{Tok}(\mathcal{G};B), \\
&
-\operatorname{Delay}(\mathcal{G};B)
\big),
\end{aligned}
\label{eq:appendix-eval-vector}
\end{equation}
where larger values are preferred in every dimension. 
\(\operatorname{Perf}\) denotes benchmark-specific task performance,
\(\operatorname{Tok}\) denotes token usage, and \(\operatorname{Delay}\) denotes
wall-clock latency. The negative signs convert efficiency objectives
into maximization dimensions.

The performance term is instantiated according to the domain of each task.
Our evolution data mainly covers ToolHop, EnvScaler-RL, and SearchQA.
For ToolHop, we combine the final-answer correctness and path-completion
metrics defined in App.~\ref{app:benchmarks}:
\begin{equation}
\begin{aligned}
\operatorname{Perf}_{\mathrm{ToolHop}}(\tau,x)
=
&
\lambda_{\mathrm{ans}}
\mathbbm{1}[\mathrm{AnsCorrect}(\tau,x)] \\
&
+\lambda_{\mathrm{path}}
\operatorname{PathScore}(\tau,x),
\end{aligned}
\end{equation}
where \(\lambda_{\mathrm{ans}}+\lambda_{\mathrm{path}}=1\), and we use
\(\lambda_{\mathrm{ans}}=\lambda_{\mathrm{path}}=0.5\) by default.

For SearchQA, we use the normalized token-level answer F1 score defined in
App.~\ref{app:benchmarks}:
\begin{equation}
\operatorname{Perf}_{\mathrm{SearchQA}}(\tau,x)
=
\operatorname{F1}(\hat{a}(\tau),a_x),
\end{equation}
where \(\hat{a}(\tau)\) is the final answer produced by trajectory \(\tau\)
and \(a_x\) is the gold answer.

For EnvScaler-RL, we follow the official environment-state verifier. Each task
scenario provides a checklist of terminal-state validation functions
\(\mathcal{C}_x=\{c^x_j\}_{j=1}^{m_x}\), where each function checks whether
one required condition is satisfied in the final environment state. Let
\(s_T(\tau,x)\) denote the terminal state reached after executing trajectory
\(\tau\). We define
\begin{equation}
\begin{aligned}
\operatorname{Perf}_{\mathrm{Env}}(\tau,x)
=
\operatorname{Done}(\tau,x)
=\\
\frac{1}{m_x}
\sum_{j=1}^{m_x}
\mathbbm{1}\!\left[
c^x_j\!\left(s_T(\tau,x)\right)=\mathrm{True}
\right].    
\end{aligned}
\end{equation}
This score measures task completion by verifying the final environment state,
rather than matching a single reference action sequence.

For a batch \(B\), each metric is averaged over all tasks in the batch:
\begin{equation}
\begin{aligned}
\operatorname{Perf}(\mathcal{G};B)
=
\frac{1}{|B|}
\sum_{x\in B}
\operatorname{Perf}_{d(x)}
\big(
\tau_x(\mathcal{G}),x
\big),
\end{aligned}
\end{equation}
where \(d(x)\) denotes the benchmark domain of task \(x\).

Given two candidate harnesses \(\mathcal{H}_a\) and \(\mathcal{H}_b\), paired with their corresponding policies to form executable systems \(\mathcal{G}_a\) and \(\mathcal{G}_b\), we say \(\mathcal{H}_a\) Pareto-dominates \(\mathcal{H}_b\) on batch \(B\) if
\begin{equation}
\begin{aligned}
\mathcal{H}_a \succ_B \mathcal{H}_b
\iff
&
\mathbf{J}(\mathcal{G}_a;B)
\succeq
\mathbf{J}(\mathcal{G}_b;B) \\
&\land
\mathbf{J}(\mathcal{G}_a;B)
\neq
\mathbf{J}(\mathcal{G}_b;B).
\end{aligned}
\end{equation}
That is, \(\mathcal{G}_a\) is no worse in every objective and strictly better in at least one objective.
During budgeted filtering, HarnessForge ranks candidates by Pareto fronts and uses the primary task performance as the tie-breaker when candidates are otherwise comparable.
The retained Pareto-competitive harnesses form the survivor set passed to policy alignment.

The same evaluator is also used to maintain the archive.
For each evaluated harness, HarnessForge stores its design summary, fault report, rollout summary, and evaluation vector \(\mathbf{J}\).
During archive-guided improvement, reference cases are retrieved based on both fault relevance and Pareto quality: the meta-agent prioritizes historical harnesses that address similar planning, action, or memory failures and lie on or near the Pareto frontier.
Thus, the evaluator provides a unified criterion for both harness selection and archive retrieval.

\subsection{Hyperparameter Configuration}
Tab.~\ref{tab:harnessforge-config} summarizes the main implementation configuration used by HarnessForge. Unless otherwise stated, the same configuration is used across benchmarks and backbones.

\begin{table}[h]
\caption{Implementation configuration of HarnessForge.}
\centering
\small
\begin{tabular}{@{}p{0.37\linewidth}p{0.57\linewidth}@{}}
\hline
\textbf{Parameter} & \textbf{Value} \\
\hline
Base reasoner & Qwen3-4B and Qwen3-8B \\
Meta-agent & GPT-5.5 \\
Evolution rounds & 3 \\
Harness proposals per round & 8 candidates \\
Filtering stages & Two-stage half-selection \\
Retained harnesses & \(|C|=2\) survivor harnesses per round \\
Harness components & Planning, Action, Memory \\
Tailoring target & Harness edits under task-interface constraints \\
Adapter type & Harness-specific LoRA adapter \\
Trainable parameters & LoRA adapters only; base model frozen \\
Adapter objective & SFT on filtered successful trajectories \\
LoRA target modules & Attention and MLP projections \\
LoRA rank / alpha / dropout & 8 / 16 / 0.05 \\
Adapter learning rate & \(2\times10^{-6}\) \\
Adapter epochs & 1 \\
\hline
\end{tabular}
\label{tab:harnessforge-config}
\end{table}

\subsection{Harness Representation and Edit Space}
\label{app:harness-edit-space}

Each harness is stored as an executable code/configuration bundle. The editable scope is restricted to the harness controller, so all candidate children can be run under the same evaluator. Tab.~\ref{tab:harness-edit-space} lists the main edit categories used by the meta-agent.

\begin{table}[t]
\caption{Executable harness representation and edit boundaries.}
\centering
\small
\begin{tabular}{@{}p{0.22\linewidth}p{0.68\linewidth}@{}}
\hline
\textbf{Component} & \textbf{Editable harness-side fields} \\
\hline
Planning \(\mathcal{P}\) & Decomposition templates, replanning triggers, verification steps, termination conditions, and controller order. \\
Action \(\mathcal{A}\) & Tool descriptions, argument schemas, routing rules, validity guards, retry policies, and role/tool orchestration. \\
Memory \(\mathcal{M}\) & State keys, write conditions, retrieval rules, summarization format, and which memory fields are exposed to the reasoner. \\
\hline
\end{tabular}

\label{tab:harness-edit-space}
\end{table}

A proposed child harness must instantiate the same controller interface as its parent and pass a lightweight validity check before rollout evaluation: required fields must be present, tool schemas must be parseable, action names must map to available tools, and memory keys referenced by planning or action code must be defined. Invalid children are discarded before budgeted selection. This representation makes the parent harness \(\rightarrow\) fault report \(\rightarrow\) child harness transition reproducible as a constrained code-editing problem rather than an unconstrained natural-language redesign.

\subsection{Meta Tailoring Operator and Prompt Protocol}
\label{app:meta-tailoring-operator}

The meta tailoring operator is implemented as a staged prompt-driven meta-agent pipeline. Instead of relying on a single monolithic generation call, the pipeline decomposes tailoring into diagnosis, improvement planning, executable harness generation, and smoke-test repair.

This staged design improves controllability and noise isolation: each stage consumes a compact, task-specific intermediate artifact rather than the full long-context evidence bundle, which reduces context drift and prevents later generation steps from being dominated by irrelevant trajectory details.
Concretely, one call diagnoses failures, one converts the diagnosis into transferable improvement directions, one writes executable harness code, and a bounded retry loop repairs implementation errors discovered by smoke tests.
The full prompt for this operation is available in our open-source codebase, and the appendix includes a representative excerpt for reference.

\subsubsection{Fault-Attribution Operation}
\label{app:fault-prompt}

The first stage localizes observed execution failures to specific harness modules. Instead of treating every wrong final answer as a generic model-reasoning failure, the meta-agent distinguishes planning, action, and memory failures.

As shown in Eq.~\eqref{eq:fault-attribution}, the fault-attribution operation produces a structured report \(\mathbf{F}_{\mathcal{H}_i}^{(r)}\) that localizes observed failures to harness modules. 
In brief, it instructs the meta-agent to analyze the current harness, rollout metrics, and representative trajectories, and to produce an evidence-grounded diagnosis rather than task-specific patches. 
The prompt requires the meta-agent to (i) reverse-engineer the implemented planning, action, memory, and wiring behavior, (ii) identify stable failure modes and their first meaningful error points, (iii) assign module ownership with concrete trajectory evidence, (iv) distinguish transferable harness weaknesses from benchmark-specific artifacts, and (v) output repair priorities, preservation constraints, and generation-ready instructions for the next tailoring stage.

\begin{tcolorbox}[notitle, sharp corners, breakable, colframe=ForestGreen, colback=white, 
       boxrule=3pt, boxsep=0.5pt, enhanced, 
       shadow={3pt}{-3pt}{0pt}{opacity=1,mygrey},
       title={Phase I Fault-Location Prompt (Partial)},]\label{box:fault-location-prompt}
       \scriptsize
       {\fontfamily{pcr}\selectfont
\begin{lstlisting}"""
# You are a Harness Failure Localization Agent
  Your goal is to analyze the execution failures of the current winner harness `{winner_harness_name}` and produce a
  **module-localized, evidence-grounded, generation-ready diagnosis of transferable harness weaknesses**.
  This is not a leaderboard summary and not a request for task-specific patches. This is a structural postmortem whose
  main purpose is to answer:
  **Which module-level capability gap is responsible for each dominant failure mode, and why is that diagnosis likely to
  transfer beyond the observed training tasks?**
  ---
  ## Harness Under Analysis

  ### Winner / Current Best Harness
  `{winner_harness_name}`

  ### Harness Snapshot
  {winner_harness_snapshot}

  ### Module Files
  `{module_files_info}`
  You may inspect the harness implementation files if tools are available. Pay special attention to:
  - `builder.py`
  - `planning_module/provider.py`
  - `action_module/provider.py`
  - `memory_module/provider.py`
  - prompt files under `planning_module/`, `action_module/`, or `memory_module/` if present
  - `Description.md` and prior reports if present
  ---
  ### PART 7: GENERATION CONSTRAINTS
  Flat bullets only.
  Each bullet must start with one of:
  - `[Planning]`
  - `[Action]`
  - `[Memory]`
  - `[Builder]`
  - `[Interface]`
  - `[Avoid]`
  - `[Preserve]`
  These should be operational constraints for the next harness generation step.
  Include at least one `[Avoid]` constraint that prevents benchmark-specific or trajectory-specific patching.
  ---
  ### PART 8: READY SIGNAL
  End with exactly:
  ```text
  READY_FOR_IMPROVEMENT_DIRECTION
  ```
  ---
  ## Critical Guidelines
  - Be evidence-first.
  - Treat trajectories as evidence for transferable harness weaknesses, not as templates for special-case fixes.
  - Do not collapse all errors into planning.
  - Treat final-answer failure as different from path-search failure.
  - Treat tool schema failure as different from reasoning failure.
  - Separate module ownership from general blame.
  - Prefer module-level capability diagnoses over small defensive patches.
  - Avoid overfitting to the current training task distribution.
  - Preserve working behaviors.
  - Make the report directly usable by the next prompt. 

"""
\end{lstlisting}
}
\end{tcolorbox}


\subsubsection{Archive-Guided Improvement}

The second stage converts the fault report into a generation-ready improvement brief.
Rather than generating the next harness from scratch, HarnessForge retrieves reference cases from the archive \(\mathcal{Z}^{(r)}\), which stores previously explored harnesses, rollout summaries, validation scores, and diagnosis reports.
Retrieval is conditioned on both fault relevance and Pareto quality: the meta-agent prioritizes historical harnesses that address similar planning, action, or memory failures and lie on or near the Pareto frontier of validation performance and execution cost.
This allows the improvement stage to reuse transferable design patterns without overfitting to narrow task-specific patches.

We retrieve the reference cases as
\begin{align}
\mathcal{S}_{\mathcal{H}_i}^{(r)}
=
\operatorname{Retrieve}
\Big(
    \mathcal{Z}^{(r)},
    \mathbf{F}_{i}^{(r)}
\Big),
\label{eq:archive-retrieval-app}
\end{align}
where \(\mathcal{S}_{\mathcal{H}_i}^{(r)}\) denotes the selected Pareto-competitive reference harnesses.
The improvement agent then produces an implementation brief:
\begin{align}
\mathbf{I}_{i}^{(r)}
=
\mathbb{R}_{\omega}
\Big(
    \mathcal{H}_i^{(r)},
    \mathbf{F}_{i}^{(r)},
    \mathcal{E}_{i}^{(r)}
\Big).
\label{eq:archive-guided-brief-app}
\end{align}

In brief, the prompt instructs the meta-agent to read the current harness, the module-localized fault report \(\mathbf{F}_{\mathcal{H}_i}^{(r)}\), and the retrieved archive examples, then produce a transferable improvement brief rather than executable code.
The brief specifies which failure modes to prioritize, which planning/action/memory or cross-module interface should own each fix, which historical design patterns should be reused or avoided, and what preservation constraints should guide the subsequent harness-generation stage.
This separates diagnosis, improvement abstraction, and code generation, reducing trajectory-level overfitting and making harness edits more transferable.
Concretely, we instantiate this operation with the following prompt:
\begin{tcolorbox}[notitle, sharp corners, breakable, colframe=ForestGreen, colback=white, 
       boxrule=3pt, boxsep=0.5pt, enhanced, 
       shadow={3pt}{-3pt}{0pt}{opacity=1,mygrey},
       title={Phase II Improvement Prompt (Partial)},]\label{box:improvement-prompt}
       \scriptsize
       {\fontfamily{pcr}\selectfont
\begin{lstlisting}"""
# You are a Harness Improvement Direction Agent
  Your goal is to read the Stage 1 module-localization report for `{winner_harness_name}`, inspect relevant harness
  examples from the harness archive / harness pool, and produce a **generation-ready improvement direction brief for
  transferable harness improvements**.
  This phase does **not** generate code.
  It decides:
  - which failure modes should be fixed first
  - which module should own each fix
  - which existing harness examples are useful as few-shot references
  - what to borrow
  - what to avoid
  - how to keep directions transferable to unseen tasks
  - what exact design direction the Stage 3 generator should follow
  ---
  ## Inputs

  ### Winner / Current Best Harness
  `{winner_harness_name}`

  ### Winner Harness Snapshot
  {winner_harness_snapshot}

  ### Stage 1 Module Localization Report
  {module_localization_report}

  ### Harness Pool / Archive Overview
  {harness_pool_overview}

  ### Candidate Harness Examples
  {harness_examples}

  ### Existing Harness Names / Identities
  {existing_harness_names}
  ---
  ## System Boundaries
  The generator can create or modify a harness candidate inside the harness factory, but it must respect the local harness structure:
  - `builder.py`
  - `__init__.py`
  - `Description.md`
  - `planning_module/provider.py`
  - `action_module/provider.py`
  - `memory_module/provider.py`
  The generator may change internal module behavior, prompts, helper functions, and module wiring as long as it remains compatible with the harness factory.

    ### PART 8: READY SIGNAL
  End with exactly:
  ```text
  READY_FOR_HARNESS_GENERATION
  ```
  ---
  ## Critical Guidelines
  - The Stage 1 localization report is authoritative.
  - Preserve Stage 1's distinction between surface trajectory evidence and transferable harness weakness.
  - Do not invent new failure modes unless the examples reveal a direct contradiction.
  - Use peer harnesses selectively rather than copying a whole architecture.
  - Prefer targeted module transfer over whole-architecture replacement.
  - Prefer reusable module capabilities over narrow fallback patches.
  - Preserve the winner's useful behaviors.
  - Make the final blueprint specific enough for code generation.

\end{lstlisting}
}
\end{tcolorbox}

\subsubsection{Refinement and Generation}

The third stage converts the improvement brief $\mathbf{I}_{i}^{(r)}$ into executable harness candidates and filters them before policy alignment.
The generator may revise the internal planning, action, memory, and wiring logic, but must preserve the dataset, evaluator, backend model, benchmark runner, and task labels.
This constraint ensures that improvements come from harness tailoring rather than task-specific shortcuts.

In brief, it instructs the meta-agent to generate a production-ready harness candidate from the current harness, the module-localized fault report $\mathbf{F}_{i}^{(r)}$, the archive-guided improvement brief $\mathbf{I}_{i}^{(r)}$, and selected reference examples.
The prompt requires the generated harness to be self-contained, importable, compatible with the expected builder interface, and organized into the required planning, action, memory, and wiring files.
It also asks the meta-agent to implement evidence-grounded module-level repairs, preserve useful parent-harness behaviors, avoid cosmetic or unrelated changes, and prevent hard-coded benchmark shortcuts.
After generation, candidates are checked by smoke tests and evaluated under a bounded rollout budget; only valid and Pareto-competitive survivors are passed to the policy-alignment stage.

Concretely, we instantiate this operation with the following prompt:
\begin{tcolorbox}[notitle, sharp corners, breakable, colframe=ForestGreen, colback=white, 
       boxrule=3pt, boxsep=0.5pt, enhanced, 
       shadow={3pt}{-3pt}{0pt}{opacity=1,mygrey},
       title={Phase III Generation Prompt (Partial)},]\label{box:generation-prompt}
       \scriptsize
       {\fontfamily{pcr}\selectfont
\begin{lstlisting}"""
  ## Stage 3 - Harness Generation

  ### Objective
  Build one creative but evolutionary, production-grade harness candidate that improves the current winner by following
  the module-localized diagnosis and the improvement direction brief.
  The generated harness must be a complete local harness bundle that can be placed under:
  `harness_factory/rounds/{target_round}/{candidate_name}/`
  Each generated harness candidate must be independent and self-contained. It may learn from the winner and selected
  examples, but it must not depend on files from sibling candidates, previous generated candidates, or mutable external
  harness directories at runtime.
  The generator should implement concrete module-level repairs, not produce another analysis report or a from-scratch
  unrelated harness.
  ---
  ### Context
  **Harness Integration**:
  A harness is composed of:
  - `builder.py` - wires the harness and exposes `build_agent_from_context(context)`
  - `__init__.py` - exports harness constants and build entry point
  - `Description.md` - explains the harness design
  - `planning_module/provider.py` - planning implementation
  - `action_module/provider.py` - action / tool-use implementation, including concrete agent collaboration / orchestration
  - `memory_module/provider.py` - memory provider or memory wrapper
  **Core Agent Constraint**:
  The benchmark runner, backend model, dataset, evaluator, and outer agent loop are fixed unless the local harness
  examples clearly show an approved extension point. The harness may evolve internally through planning behavior,
  action behavior, memory guidance, and simple interfaces between them.
  **Module Ownership**:
  - Planning owns decomposition, plan topology, progress state, verification questions, evidence-chain intent,
    replan triggers/timing, milestone gates, and commit criteria.
  - Action owns concrete agent collaboration / orchestration, agent topology, tool selection, tool-call arguments,
    schema preflight/repair, observation handling, retry/stop policy, aggregation/arbitration, and final-answer submission.
  - Memory owns compact reusable guidance, prior-case reminders, memory encoding, storage scope/format,
    read/retrieval timing, routing, update/write policy, and ingestion.
  - Builder owns wiring, constants, metadata, provider connection, and harness import compatibility.
  - Interfaces should be simple and explicit. Use them only when one module must pass state or guidance to another.
  ---
  ## Final Checklist
  Use flat bullets only.
  Include:
  - unique candidate name
  - candidate is independent and self-contained
  - required files present
  - builder exports correct entry point
  - module constants consistent
  - harness is intended to be executable without manual repair
  - imports and provider exports are complete
  - Stage 1 top failures addressed
  - Stage 2 borrow / avoid constraints followed
  - no hard-coded answers
  - no whole-harness copy
  - no unnecessary architecture expansion
  ---
  ### Final Reminder
  The generated harness should be ambitious in completeness and creative in local mechanisms, but conservative in overall
  architecture.
  It should feel like a precise module-level evolution of the current winner, not a new unrelated system.

\end{lstlisting}
}
\end{tcolorbox}

\subsubsection{Retry Mechanism}

Generated harnesses may contain implementation errors even when their high-level design is useful.
Before rollout validation, we apply a lightweight smoke-test for importability, builder-interface compatibility, provider signatures, tool-call format validity, and execution on diagnostic examples.
Failed candidates are returned to the meta-agent with the error trace and intended improvement brief for repair.
We allow at most \(N_{\mathrm{retry}}=3\) repair attempts; candidates that still fail are discarded before validation. This prevents invalid harnesses from consuming rollout budget, and the surviving executable harnesses are passed to harness-conditioned policy alignment.

\subsubsection{Filter and Archive Update}
\label{app:archive_update}

HarnessForge uses a budgeted half-selection protocol to filter generated harness candidates before policy alignment.
Each evolution round uses an evolution batch of about \(1.2\mathrm{K}\) tasks.
Instead of evaluating every candidate on the full batch, we apply staged filtering on progressively consumed subsets.
In our default setting, each filtering stage evaluates candidates on \(200\) tasks.

Let \(\mathcal{C}_{0}^{(r)}\) denote the pooled candidate set generated at round \(r\).
At filtering stage \(t\), each candidate harness \(\mathcal{H}\in\mathcal{C}_{t-1}^{(r)}\) is paired with its corresponding policy to form an executable system \(\mathcal{G}_{\mathcal{H}}\), and is evaluated on a fresh filtering subset \(B_{r,t}\) with \(|B_{r,t}|=200\).
We compute the multi-objective evaluation vector \(\mathbf{J}(\mathcal{G}_{\mathcal{H}};B_{r,t})\), rank candidates by Pareto dominance with task performance as the primary tie-breaker, and retain the top half:
\begin{equation}
\begin{aligned}
\mathcal{C}_{t}^{(r)}
=
\operatorname{Half}_{\mathrm{Pareto}}
\Big(
\mathcal{C}_{t-1}^{(r)},
\{
\mathbf{J}(\mathcal{G}_{\mathcal{H}};B_{r,t})
\}_{\mathcal{H}\in\mathcal{C}_{t-1}^{(r)}}
\Big).
\end{aligned}
\end{equation}

We use two half-selection stages by default.
Thus, starting from eight generated candidates, the first \(200\)-task filtering stage retains four harnesses, and the second \(200\)-task filtering stage retains two survivor harnesses.
The remaining tasks in the round batch are then used to finish rollout evaluation for the survivor harnesses.
These survivor rollouts are reused for harness-conditioned policy alignment, so the policy-alignment stage does not require a separate data-collection phase.

After each filtering stage, HarnessForge updates the archive \(\mathcal{Z}^{(r)}\).
For every evaluated harness, the archive stores its design summary, parent lineage, fault report, improvement brief, rollout summary, selection status, and evaluation vector \(\mathbf{J}\).
The updated archive is used in later rounds for Pareto-aware retrieval: archive cases are selected not only by fault relevance, but also by whether they lie on or near the Pareto frontier of performance and efficiency.
This makes the filtering process both budget-efficient and reproducible, while also turning past evaluations into reusable evidence for subsequent harness tailoring.

\section{Policy Alignment Details}
\label{app:policy-alignment-instantiations}

The policy side of HarnessForge is always conditioned on the selected harness and its lineage. This appendix separates the alignment protocol from the baseline-training configuration in App.~\ref{app:baseline-config}.

\subsection{Policy Lineage and Adapter Operation}
\label{app:policy-lineage}

HarnessForge keeps a lineage for each survivor pair. When multiple survivor harnesses descend from the same parent pair, each child receives an independent copy of the parent policy state. The copied policy is then adapted separately using trajectories collected under the corresponding child harness, so sibling survivor pairs do not share trainable adapter parameters after branching.

Operationally, we first materialize the parent lineage policy as the initialization for the child pair. The parent adapter state is folded into the child initialization, and a new harness-specific LoRA adapter is attached and trained on successful trajectories collected under the child harness. We write this update as
\begin{equation}
    \delta_k^{(r+1)}
    =
    \operatorname{Mat}\!\left(\delta_{\mathrm{parent}(k)}^{(r)}\right)
    \oplus
    \Delta\delta_k^{(r+1)},
\end{equation}
where \(\operatorname{Mat}(\cdot)\) denotes materializing the parent lineage policy for the child branch, \(\oplus\) denotes attaching the new round-specific adapter to that materialized initialization, and \(\Delta\delta_k^{(r+1)}\) is trained independently for harness \(\mathcal{H}_k^{(r+1)}\). This convention makes pair lineage explicit while preserving the matched harness--policy interpretation used in the compatibility analysis.

\subsection{Trajectory Curation and Success Filtering}

For each survivor harness, HarnessForge reuses rollout traces already collected during staged harness filtering. A trajectory is retained for policy alignment only when it satisfies the task-success signal used by the corresponding benchmark evaluator. The retained trajectories are then decomposed into step-level decision pairs over the active harness interface, observation history, memory state, available actions, and next behavior. This reuse keeps the default SFT instantiation rollout-neutral with respect to the harness-selection budget.

\subsection{Objective Instantiations}

\paragraph{Training-objective agnostic alignment.}
The policy-alignment step in HarnessForge is not tied to a specific training objective. Given a harness \(\mathcal{H}\) and a set of harness-conditioned trajectories collected under \(\mathcal{H}\), the goal is to update the adapter so that the reasoner can execute reliably within the interface defined by that harness. We therefore view this step as a harness-conditioned policy-alignment operator rather than an SFT-specific operator. This operator can be instantiated with supervised fine-tuning, preference optimization, or reinforcement learning objectives. In this sense, HarnessForge defines the harness-conditioned adaptation problem, while the specific optimizer determines how the policy side is updated.

\paragraph{Default policy-evolution objective.}
In the main experiments, HarnessForge instantiates policy evolution with supervised fine-tuning. We use SFT for two practical reasons. First, it preserves the rollout budget: the successful trajectories used for policy learning are already collected during budgeted harness selection, so the policy step does not require a new environment-interaction phase,
\begin{equation}
    \mathcal{T}_k^{+}\subseteq\mathcal{T}_k^{(r+1)}.
\label{eq:sft-rollout-budget}
\end{equation}
Second, SFT gives a direct step-level supervision signal. Each retained successful rollout is decomposed into decision-level pairs,
\begin{equation}
    \mathcal{D}_{\mathcal{H}_k}
    =
    \{(z_t,y_t)\mid \tau\in\mathcal{T}_k^{+},~t=1,\ldots,|\tau|\}.
\label{eq:sft-step-dataset}
\end{equation}
Here \(z_t\) contains the task, active harness specification, observation history, memory state, and available actions; \(y_t\) is the demonstrated next behavior under that harness, such as a reasoning step, tool call, memory write, or final answer. The corresponding objective is
\begin{equation}
    \mathcal{L}_{\mathrm{sft}}
    =
    -\sum_{(z_t,y_t)\in\mathcal{D}_{\mathcal{H}_k}}
    \log p_{\mathcal{R}^{(r)}_{\delta_k}\oplus\Delta\delta}(y_t\mid z_t).
\label{eq:appendix-sft-loss}
\end{equation}
Thus, SFT does not optimize a separate scalar reward; it teacher-forces the executor to imitate successful harness-conditioned decisions. This makes it a natural and rollout-efficient objective for learning harness-specific execution behaviors, including action formatting, tool-use discipline, memory utilization, verification habits, and termination control.

\paragraph{Policy-alignment objective agnosticism.}
The policy-alignment step in HarnessForge is defined at the framework level rather than being tied to one particular optimization objective. Its role is to adapt the executor to the selected harness, so the same operator can be instantiated with SFT, preference-style objectives, or RL-style objectives. In this view, different objectives mainly provide different ways of converting harness-conditioned execution evidence into an updated policy: SFT imitates curated successful trajectories, preference-style training contrasts stronger and weaker executions, and RL-style training directly optimizes reward signals under the current harness.

This design makes the choice of policy optimizer a practical tradeoff rather than a methodological constraint. As shown in Tab.~\ref{tab:training-method-agnosticism}, GRPO and RLOO can bring additional improvements in several settings, suggesting that HarnessForge still has headroom when stronger policy optimization and larger interaction budgets are available. However, these methods typically require extra rollout sampling for candidate generation, preference construction, or reward estimation. By contrast, SFT can reuse trajectories already collected during harness evolution, making it a more budget-efficient default. We therefore use SFT in the main experiments as a conservative tradeoff between performance and rollout cost, while treating preference- and RL-style objectives as higher-budget instantiations of the same alignment step.

\section{Baselines and Fairness Protocol}
\label{app:baseline-config}

\subsection{Shared Fairness Protocol}
For fair comparison, training-style baselines such as SFT, GRPO, and RLOO are paired with the same fixed base scaffold and trained on the same 3.8K trajectory pool used by HarnessForge for trajectory curation and policy alignment. These baselines therefore isolate the effect of policy optimization under a fixed execution scaffold, without performing harness evolution. Meta-agent-based harness-search baselines use the same GPT-5.5 meta-agent backend as HarnessForge. Following prior agent-search work, rollout budget counts executable environment/model rollouts, while meta-agent generation cost is treated as implementation overhead and is not included in rollout counts. The same convention is applied to HarnessForge and all meta-agent-based harness-search baselines.

\subsection{Search-Style Baselines}
\label{app:search-style-baselines}

Search-style harness baselines use the same frozen base reasoners as HarnessForge and optimize only the external execution structure. The executor is served through an OpenAI-compatible local vLLM endpoint with deterministic decoding (\(\text{temperature}=0\), \(\text{top-p}=1\)), Qwen3 thinking disabled, and a 300-second request timeout. The held-out test split is never used during search. TMDB and API-Bank are used only for transfer evaluation rather than task-specific evolution. For TMDB and API-Bank, all methods start from the ToolHop-evolved harness or workflow and are evaluated under the target API interface without using TMDB or API-Bank held-out test examples. Tab.~\ref{tab:search-baseline-config} summarizes the executable search budgets used in our local baseline ports.

\begin{table*}[t]
\caption{Executable configuration of search-style harness baselines. Validation statistics are used only for model or workflow selection; held-out test scores are reported in Tab.~\ref{tab:main-result}.}
\centering
\scriptsize
\setlength{\tabcolsep}{2.5pt}
\renewcommand{\arraystretch}{1.08}
\begin{tabularx}{\textwidth}{@{}p{0.10\textwidth}p{0.22\textwidth}p{0.23\textwidth}p{0.20\textwidth}p{0.20\textwidth}@{}}
\toprule
\textbf{Method} & 
\textbf{Evolution Data} & 
\textbf{Search Budget / Max Rounds} & 
\textbf{Validation Selection Criterion} & 
\textbf{Task Interface / Executor} \\
\midrule

ADAS 
& ToolHop: 800 online-dev examples; SearchQA: 1000 offline validation examples. 
& Target 30 valid agent generations; up to 3 debug attempts and 3 meta retries; invalid generations skipped. 
& Select by ToolHop hybrid validation score or SearchQA validation exact-match score. 
& ToolHop closed-mode tool calls; SearchQA E5 retrieval, top-\(3\), max 4 searches, 6000-char observations; frozen executor. \\

AFlow 
& ToolHop: 800 online-dev examples; SearchQA: 1000 offline validation examples. 
& Top-\(4\) workflow sampling per round; one validation pass per workflow; SearchQA search capped at 20 rounds. 
& Select the workflow with the best validation score on the corresponding task split. 
& Code-represented workflows with ToolHop/SearchQA tool-call operators and no executor training. \\

AgentSquare 
& ToolHop: 800 online-dev examples; SearchQA: 1000 offline validation examples. 
& Enumerates 1050 modular combinations from planning, reasoning, tool-use, and memory modules; target 30 valid generations. 
& Select the best valid modular agent by validation performance. 
& Same adapters as ADAS; invalid generated modules skipped; retrieval settings matched to ADAS. \\

MaAS 
& ToolHop: 800 online-dev examples; SearchQA: 1000 offline validation examples. 
& Adam controller search with lr \(0.01\), batch size 4, sample size 4; up to 9 recorded training rounds. 
& Select by controller validation objective on the development split. 
& Operator-controller search over 5--6 operators; no deterministic path repairs. \\

MermaidFlow 
& ToolHop: 800 online-dev examples; SearchQA: 1000 offline validation examples. 
& Elite workflow sampling with Mermaid-code validation; recombination after round 4 with probability \(0.1\); search capped by recorded evolution rounds. 
& Select the best executable Mermaid workflow by validation performance. 
& Mermaid graph plus executable code; SearchQA E5 retrieval, top-\(3\), max 4 searches; deterministic executor, no policy tuning. \\

\bottomrule
\end{tabularx}
\label{tab:search-baseline-config}
\end{table*}

\paragraph{Baseline-specific notes.}
\textbf{ADAS} is used as a free-form executable agent-code search baseline; we port generated agents to the ToolHop and SearchQA runtimes and discard invalid or schema-breaking generations before held-out evaluation.
\textbf{AFlow} searches Python workflow programs with task-specific tool-call operators, so it tests whether workflow-level code evolution alone can match paired harness--policy adaptation.
\textbf{AgentSquare} provides a structured modular-search counterpart, combining planning, reasoning, tool-use, and memory modules under the same task adapters.
\textbf{MaAS} searches an operator controller for multi-agent execution rather than a single explicit workflow; deterministic path-repair shortcuts are disabled in our port.
\textbf{MermaidFlow} follows the AFlow-style workflow search protocol but constrains candidates to a Mermaid graph plus executable code representation.
Across all search-style baselines, the executor policy is not trained; only the external harness or workflow structure is selected.

\paragraph{Transfer setting for TMDB and API-Bank.}
TMDB and API-Bank are used only for transfer evaluation rather than task-specific harness search, because they do not provide dedicated evolution or training splits in our setting. For these two benchmarks, we start from the workflow or harness evolved on ToolHop and adapt it to the new task interface by replacing the tool APIs, environment wrappers, action schemas, and benchmark-specific execution constraints. No TMDB or API-Bank held-out test examples are used during this adaptation. This setting evaluates whether a harness discovered on ToolHop encodes reusable agent-system execution patterns that can be re-instantiated under new tools and environments.

\subsection{Training-Style Baselines}
\label{app:training-style-baselines}

\begin{figure}[t]
  \centering
  \begin{minipage}[t]{0.48\textwidth}
    \centering
    \includegraphics[width=\linewidth]{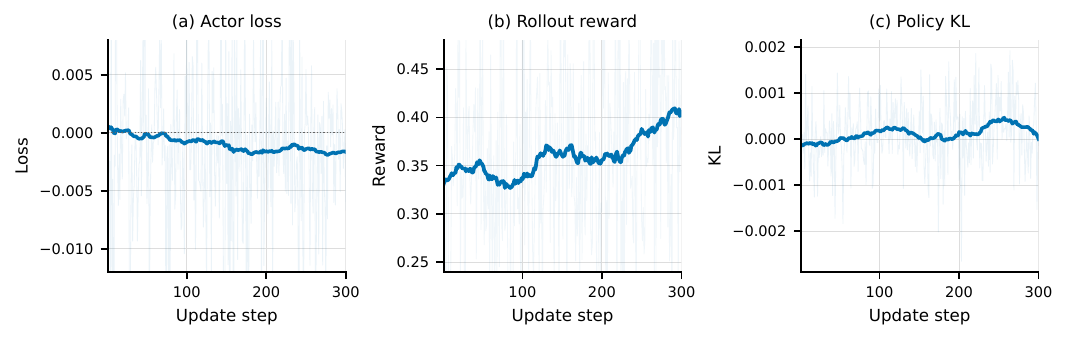}
    
    \vspace{-0.04in}
    {\small (a) RLOO training diagnostics}
  \end{minipage}
  \hfill
  \begin{minipage}[t]{0.48\textwidth}
    \centering
    \includegraphics[width=\linewidth]{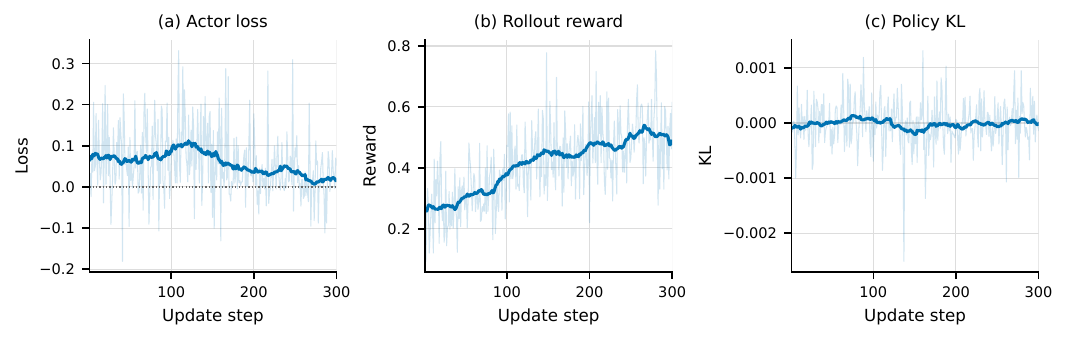}
    
    \vspace{-0.04in}
    {\small (b) GRPO training diagnostics}
  \end{minipage}
  \caption{
  Online optimization diagnostics for the RLOO and GRPO training-style baselines over the first 300 update steps. 
  We report actor loss, rollout reward, and policy KL from the training learner. 
  Curves are smoothed for readability; these traces are used only as training diagnostics, while final comparisons are based on blind-test evaluation.
  }
  \label{fig:rl-training-trajectories}
\end{figure}

Training-style baselines keep the harness fixed and update only the model-side executor.
Unless otherwise stated, the base reasoner is frozen and only LoRA adapters are updated.
Tab.~\ref{tab:grpo-config} summarizes the shared adapter-training configuration, while Fig.~\ref{fig:rl-training-trajectories} reports online optimization traces for the RL-style baselines. Final comparisons are based on held-out test evaluation in Tab.~\ref{tab:main-result}.

\paragraph{SFT.}
SFT trains LoRA adapters with multi-turn trajectories collected under the selected fixed-harness interface. We use rejection sampling to construct the training set: only trajectories with valid final answers and positive task feedback are retained, while failed executions, malformed action traces, and trajectories without valid answers are discarded. Given a curated trajectory dataset \(\mathcal{D}_{\mathrm{sft}}=\{(x_i,y_i)\}_{i=1}^{|\mathcal{D}_{\mathrm{sft}}|}\), where \(x_i\) denotes the task prompt together with the harness-conditioned interaction history and \(y_i\) denotes the target assistant action/response sequence, we optimize the token-level negative log-likelihood
\begin{equation}
\begin{aligned}
\mathcal{L}_{\mathrm{SFT}}(\delta)
=-\mathbb{E}_{(x,y)\sim \mathcal{D}_{\mathrm{sft}}}\\
\left[
\frac{1}{|y|}
\sum_{t=1}^{|y|}
\log \pi_{\theta_0+\delta}(y_t \mid x,y_{<t})
\right],   
\end{aligned}
\end{equation}

where the backbone parameters \(\theta_0\) are frozen and only the LoRA adapter \(\delta\) is updated. This instantiation is rollout-budget efficient because it reuses successful trajectories already produced during harness evolution rather than requiring an additional exploration stage.

\paragraph{Reward Design.}
All RL-style baselines use the same function-based verifier reward rather than a learned reward model.
The rollout worker executes the complete scaffold-conditioned tool trajectory and assigns a scalar reward from the final trace.
For ToolHop, the reward combines final-answer correctness and a path score that measures whether intermediate tool observations cover the annotated subtask answers, with equal weights \(0.5/0.5\).
Missing or malformed final answers incur a \(-0.1\) penalty, and non-final tool calls beyond four are penalized by \(0.01\) per call; the resulting score is clipped to \([-0.1,1]\).
For SearchQA, exact-match final answers receive full credit, substring matches receive \(0.5\) partial credit, and responses without an extractable answer receive zero.
For EnvScaler-style stateful tasks, the terminal reward is the fraction of task postcondition checks satisfied after the agent calls the completion action.
GRPO and RLOO share this reward function and the same terminal environment feedback, so their comparison isolates the advantage estimator rather than reward design.

\paragraph{GRPO.}
GRPO normalizes group-relative advantages within the \(N=4\) completions for each prompt. 
KL is not added to the reward, instead, both methods use an actor-side low-variance KL loss with coefficient \(10^{-3}\).

\paragraph{RLOO.}
RLOO uses the same on-policy scaffold agent loop as GRPO. For each prompt, the rollout worker samples \(N=4\) trajectories. The advantage of a trajectory is computed against the mean reward of the other trajectories in the same prompt group, giving a leave-one-out baseline without training a separate value critic. We keep the clipped actor objective, token-mean loss aggregation, entropy coefficient, KL loss, and rollout budget identical to GRPO so that the comparison isolates the advantage estimator.

\begin{table}[t]
\caption{Training configuration for training-style baselines.}
\centering
\small
\begin{tabular}{@{}p{0.37\linewidth}p{0.57\linewidth}@{}}
\hline
\textbf{Parameter} & \textbf{Value} \\
\hline
Trainable parameters & LoRA adapters only \\
LoRA target modules & Attention and MLP projections \\
LoRA rank & 8 \\
LoRA alpha & 16 \\
LoRA dropout & 0.05 \\
Epoch & 1.0 \\
Optimizer & AdamW with bf16 training \\
Learning rate & \(2\times10^{-6}\) \\
LR sweep & \(\{5\times10^{-6},1\times10^{-5},2\times10^{-5}\}\) \\
Scheduler & Cosine decay, \(3\%\) warmup \\
KL coefficient & 0.01 \\
Clip range & 0.2 \\
Reward & Success, validity, tool call counts \\
\hline
GRPO $rollout_n$ & 4 rollouts per prompt \\
RLOO $rollout_n$ & 4 rollouts per prompt \\
\hline
\end{tabular}
\label{tab:grpo-config}
\end{table}

\begin{figure*}[!t]
  \centering
  \includegraphics[width=\textwidth]{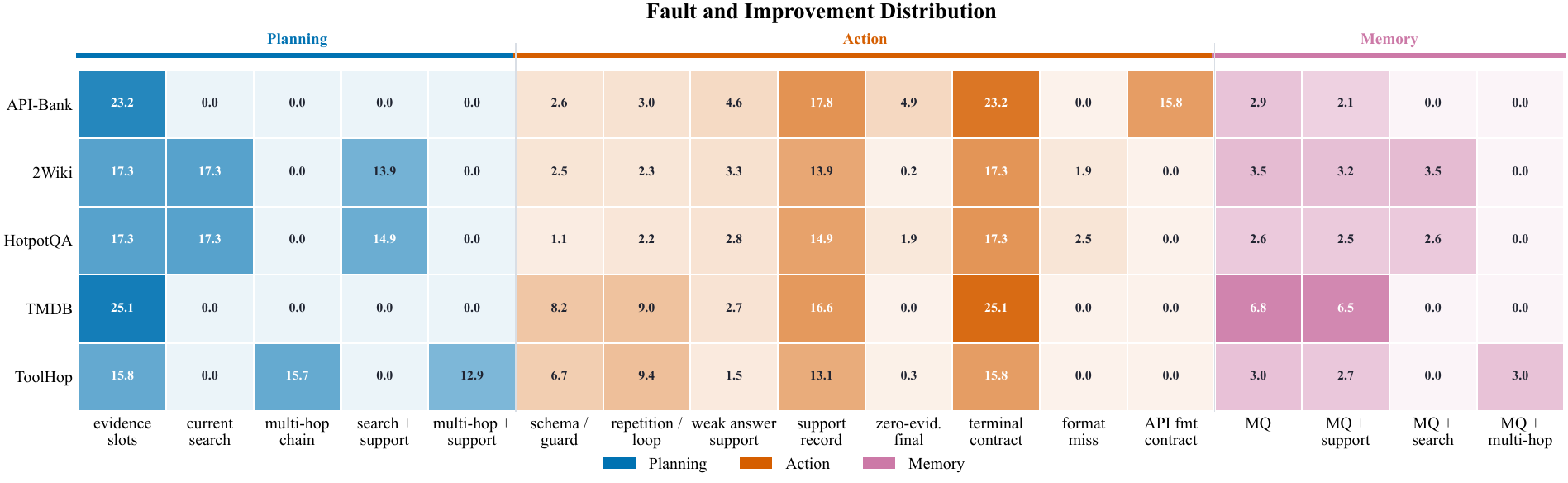}
  \caption{
  Fault and improvement-signal distribution across benchmarks.
  Each row is normalized within a benchmark, and columns group repair signals into
  planning-, action-, and memory-related categories.
  }
  \label{fig:module_repair_heatmap}
  \vskip -0.2in
\end{figure*}

\section{Reproducibility Artifacts}

\paragraph{Reproducibility of Meta-Agent Harness Evolution.}
HarnessForge uses a strong meta-agent for harness evolution, but implements it as a staged operator with fixed input-output contracts rather than an unconstrained generation call. Each stage takes structured inputs, including the parent harness, rollout evidence, evaluation summaries, and archive records, and produces schema-controlled artifacts such as fault reports, evolution manifests, harness-candidate manifests, and smoke-test logs. These artifacts record the diagnosed failure modes, responsible modules, supporting evidence, edited components, repair priorities, and expected behavioral changes, making the evolution process auditable rather than opaque.

We support reproducibility at two levels. For result-level replay, we release the final selected harness--policy pairs, including evolved harnesses, adapter checkpoints and configurations, benchmark wrappers, evaluation scripts, and split identifiers. This allows the reported systems to be evaluated without re-running meta-agent evolution. For process-level auditability, we release the prompts, operator schemas, generated reports, candidate manifests, smoke-test outcomes, and survivor-selection logs used during evolution. These records expose how each harness was generated, what failure it aimed to repair, and why it was retained.

Because meta-agent outputs may vary across model versions and providers, we do not assume exact regeneration of every intermediate candidate. Instead, our artifact release separates replaying the final executable systems from re-instantiating the evolution protocol. The released schemas and stage-wise contracts allow future work to replace the proprietary meta-agent with alternative closed- or open-source models under the same evolution interface and selection procedure.

\section{Additional Results and Analysis}

\subsection{Adaptation Necessity Analysis}
\label{app:compatibility_matrix}

\begin{figure}[!h]
  \centering
  \includegraphics[width=0.98\columnwidth]{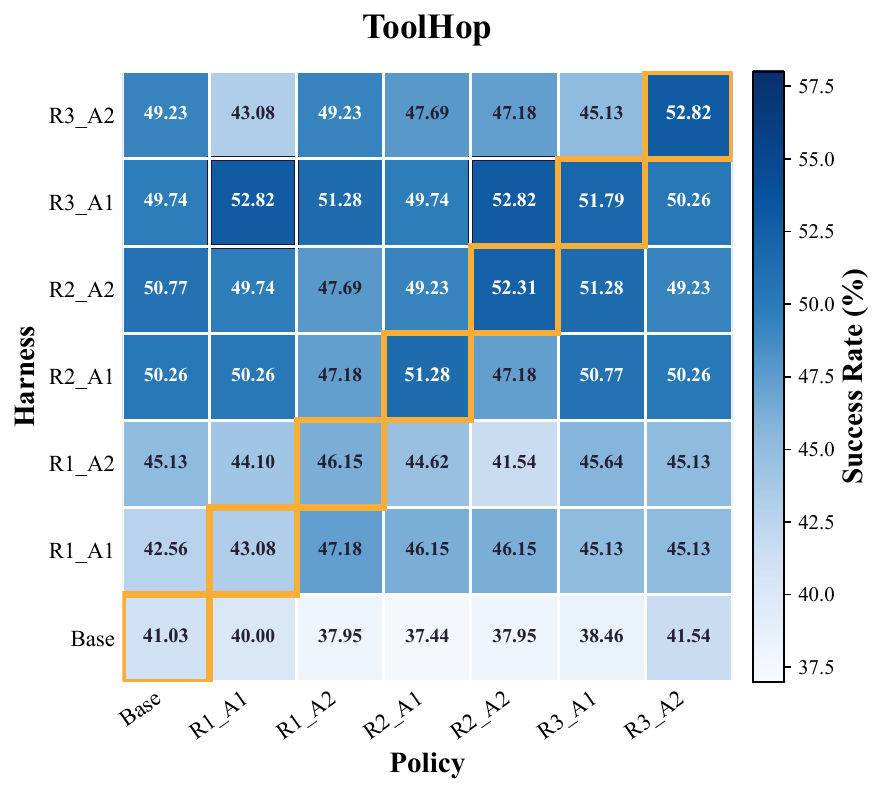}
  \vskip -0.1in
  \caption{Harness--policy compatibility matrix on ToolHop. Rows denote evolved harnesses and columns denote evolved policies across co-evolution rounds.}
  \label{fig:compatibility_matrix_toolhop}
  \vskip -0.15in
\end{figure}

Figs.~\ref{fig:compatibility_matrix_api-bank} and Figs.~\ref{fig:compatibility_matrix_toolhop} provide full harness--policy compatibility matrices on API-Bank and ToolHop.
Rows correspond to evolved harnesses, columns correspond to evolved policies, and each cell evaluates one cross-combination of a harness and a policy.
The diagonal entries therefore represent the matched harness--policy pairs produced by HarnessForge across co-evolution rounds.

The matrices show that matched pairs generally achieve stronger performance than mismatched combinations.
Along the diagonal, performance improves as evolution proceeds, indicating that gains accumulate through progressive harness--policy co-evolution rather than from a single late-stage component.
Off-diagonal entries reveal the compatibility gap: later policies do not always transfer cleanly to earlier or non-corresponding harnesses, and strong harnesses may underperform when paired with misaligned policies.
This pattern supports our main claim that HarnessForge improves agent systems by evolving compatible harness--policy pairs, rather than independently optimizing reusable harnesses or universally stronger policies.

\subsection{Module-Level Repair Statistics}
\label{app:module_repair_statistics}

Fig.~\ref{fig:module_repair_heatmap} shows that harness failures are
module-specific and benchmark-dependent. Action-side signals dominate on
API-heavy benchmarks such as API-Bank and TMDB, where common failures include
invalid API formats, schema mismatches, missing guards, repeated actions, and
incorrect endpoint or tool selection. These tasks therefore benefit most from
action-layer repairs such as stricter format contracts, schema preflight, and
loop prevention.

Planning-side signals are more visible on retrieval-heavy and multi-hop
benchmarks. The SearchQA variants mainly require current-query repair and
stronger coupling between search actions and supporting evidence, while ToolHop
more often requires preserving multi-hop tool chains and grounding final answers
in the accumulated evidence. These patterns indicate that planning repairs are
most useful when success depends on maintaining the right intermediate intent
across several reasoning or retrieval steps.

Memory-related signals are smaller but still meaningful. They typically appear
as co-repairs with planning or action modules, suggesting that memory quarantine
acts as a stabilizing layer rather than an isolated repair target. It helps
prevent stale or irrelevant traces from disrupting otherwise valid plans and
tool actions. Overall, the heatmap suggests a clear division of repair pressure:
API tasks stress action reliability, search tasks stress query planning and
evidence coupling, multi-hop tool-use tasks stress path preservation, and memory
repairs provide cross-cutting execution stability.

\subsection{Case Study}
\label{app:case_study}

\begin{figure*}[t]
  \centering
  \includegraphics[width=\textwidth]{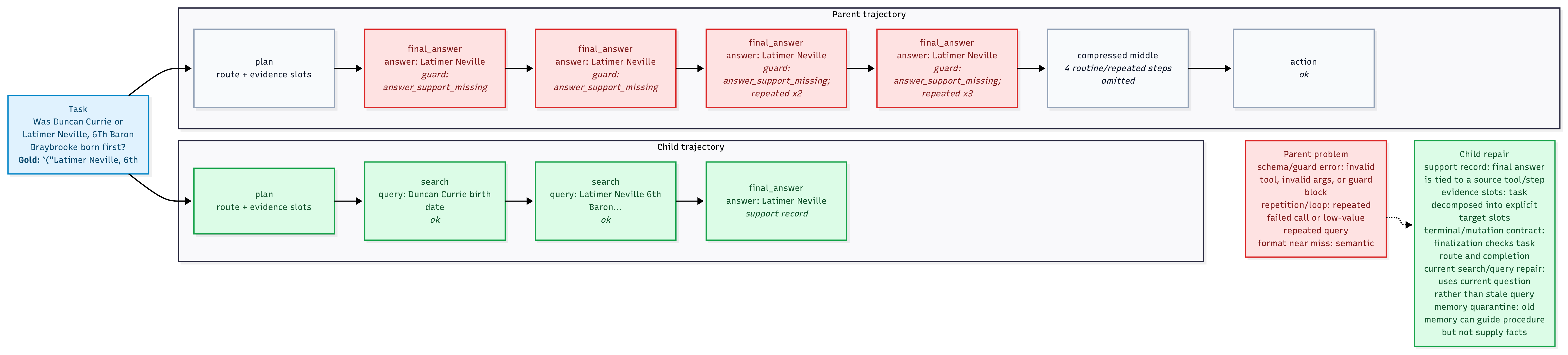}
  \caption{
  Case Study 1: evidence-supported finalization for multi-hop comparison.
  The child harness replaces repeated unsupported finalization with entity-level
  evidence slots and support-record verification.
  }
  \label{fig:case_multihop_comparison}
  \vspace{-0.2in}
\end{figure*}

\begin{figure*}[t]
  \centering
  \includegraphics[width=\textwidth]{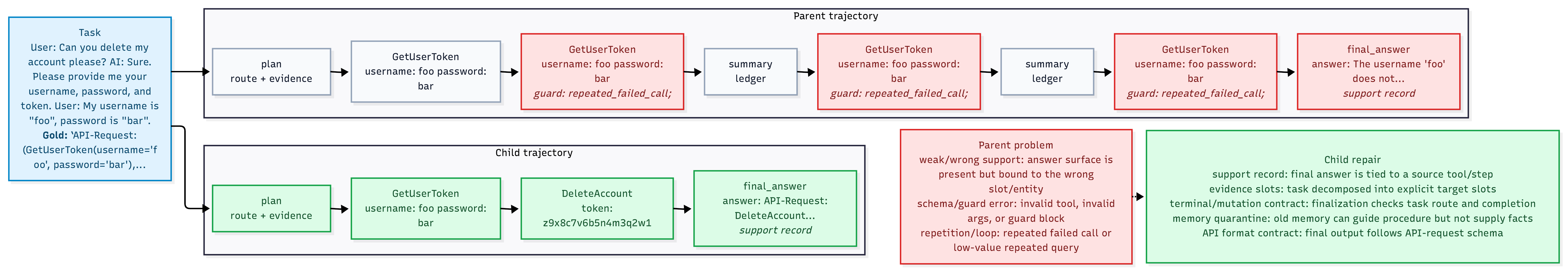}
  \caption{
  Case Study 2: API-contract repair for account deletion.
  The child harness follows the required API route and emits the final answer in
  the expected API-request format.
  }
  \label{fig:case_api_account_deletion}
  \vspace{-0.2in}
\end{figure*}

\begin{figure*}[t]
  \centering
  \includegraphics[width=\textwidth]{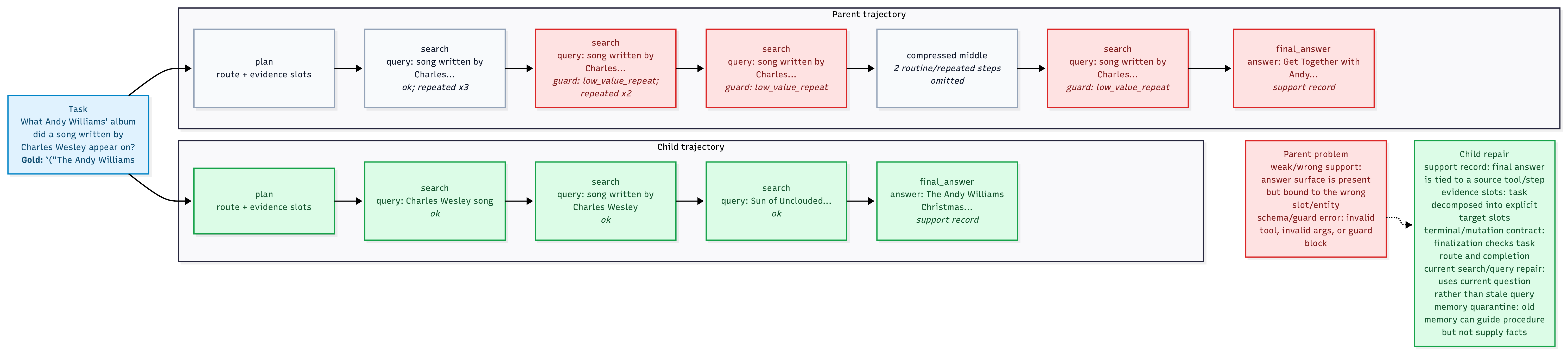}
  \caption{
  Case Study 3: current-query repair for retrieval-heavy multi-hop QA.
  The child harness replaces stale repeated searches with question-grounded
  queries and support-record finalization.
  }
  \label{fig:case_searchqa_query_repair}
  \vspace{-0.2in}
\end{figure*}

\begin{figure*}[t]
  \centering
  \includegraphics[width=\textwidth]{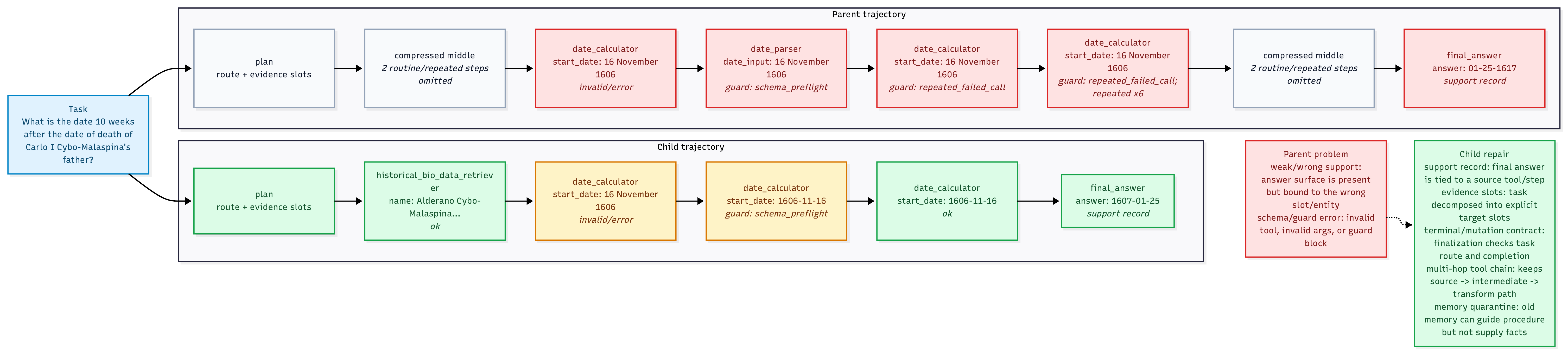}
  \caption{
  Case Study 4: schema-guarded multi-hop tool-chain repair.
  The child harness preserves the source--intermediate--transform path while
  repairing invalid date-tool arguments.
  }
  \label{fig:case_toolhop_date_chain}
  \vspace{-0.2in}
\end{figure*}

\begin{figure*}[t]
  \centering
  \includegraphics[width=\textwidth]{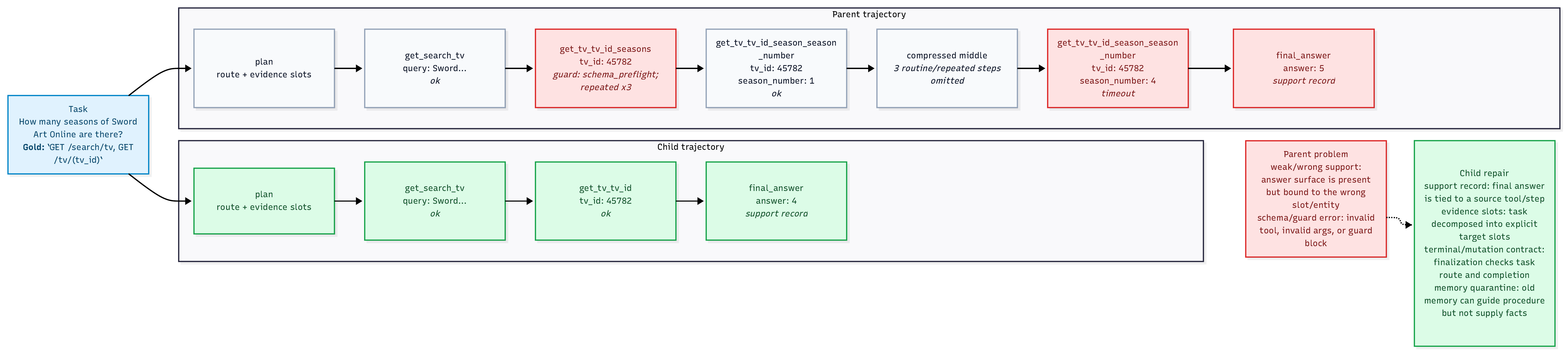}
  \caption{
  Case Study 5: endpoint-routing repair in TMDB-style API use.
  The child harness routes to the correct TV-detail endpoint and avoids
  unnecessary repeated season-level calls.
  }
  \label{fig:case_tmdb_endpoint}
  \vspace{-0.2in}
\end{figure*}

To complement the aggregate repair statistics in
Fig.~\ref{fig:module_repair_heatmap}, we present five representative
parent--child trajectory comparisons in
Figs.~\ref{fig:case_multihop_comparison}--\ref{fig:case_tmdb_endpoint}.
Each case shows how HarnessForge turns a recurring trajectory-level failure
into a reusable repair over planning, action, or memory.

\paragraph{Case 1: Multi-hop comparison.}
Fig.~\ref{fig:case_multihop_comparison} shows that the child harness replaces
repeated unsupported finalization with entity-level evidence slots and
support-record verification.

\paragraph{Case 2: Structured API execution.}
Fig.~\ref{fig:case_api_account_deletion} illustrates API-route repair: the child
harness follows the required authentication--deletion sequence and enforces the
expected API-request output format.

\paragraph{Case 3: Retrieval-heavy multi-hop QA.}
Fig.~\ref{fig:case_searchqa_query_repair} shows current-query repair, where the
child harness avoids stale repeated searches and grounds the final answer in
question-specific evidence.

\paragraph{Case 4: Multi-hop tool-chain execution.}
Fig.~\ref{fig:case_toolhop_date_chain} demonstrates schema-guarded tool use:
the child harness preserves the source--intermediate--transform chain while
normalizing invalid date-tool arguments.

\paragraph{Case 5: REST-style endpoint routing.}
Fig.~\ref{fig:case_tmdb_endpoint} shows endpoint-routing repair, where the child
harness avoids unnecessary season-level calls and retrieves the answer from the
correct TV-detail endpoint.

Overall, these cases show that HarnessForge improves execution behavior by
modifying reusable interfaces, such as evidence slots, schema guards, support
records, API contracts, and memory quarantine, rather than applying
task-specific answer patches.

\end{document}